%% file: main.tex
\documentclass[10pt,journal,compsoc]{IEEEtran}

\usepackage{graphicx}
\usepackage{comment}
\usepackage{amsmath,amssymb,amsfonts} 
\usepackage{color}
\usepackage{times}
\usepackage{epsfig}
\usepackage{enumitem}
\usepackage{booktabs}
\usepackage{multirow}
\usepackage{graphicx}
\usepackage{caption}
\usepackage{mathtools}
\usepackage{url}
\usepackage{subfiles}
\usepackage{bbold}
\usepackage[dvipsnames]{xcolor}
\usepackage{floatrow}
\usepackage{microtype}
\usepackage{booktabs}
\usepackage{colortbl}
\usepackage{diagbox}
\usepackage[ruled,vlined]{algorithm2e}
\usepackage{soul}

\usepackage{hyperref}
\hypersetup{
    colorlinks=true,
    linkcolor=black,
    filecolor=magenta,      
    urlcolor=blue,
}
\DeclareMathOperator*{\argmin}{arg\,min}

\def\eqnvspace{{\vspace{-0.5mm}}}
\def\figvspace{{\vspace{-3mm}}}
\newcommand{\Paragraph}[1]{\vspace{-0mm} \noindent \textbf{#1} \hspace{0mm}}
\newcommand{\Section}[1]{\vspace{-0mm} \section{#1} \vspace{-0mm}}
\newcommand{\SubSection}[1]{\vspace{-0mm} \subsection{#1} \vspace{-0mm}}
\newcommand{\SubSubSection}[1]{\vspace{-0mm} \subsubsection{#1} \vspace{-0mm}}

\definecolor{XL_color}{rgb}{0.858, 0.188, 0.478}

\definecolor{YJ_color}{rgb}{0.1, 0.188, 0.878}

\ifCLASSOPTIONcompsoc
  \usepackage[nocompress]{cite}
\else
  \usepackage{cite}
\fi

\begin{document}
\title{Physics-Guided Spoof Trace Disentanglement for Generic Face Anti-Spoofing}

\author{Yaojie~Liu, and~Xiaoming~Liu,~\IEEEmembership{Member,~IEEE}
\IEEEcompsocitemizethanks{\IEEEcompsocthanksitem 
Y. Liu and X. Liu are with the Department of Computer Science and Engineering, Michigan State University, East Lansing, MI 48824,  USA. \protect\\ Email: \{liuyaoj1,liuxm\}@msu.edu.}
}

\markboth{Journal of \LaTeX\ Class Files,~Vol.~14, No.~8, August~2015}%
{Shell \MakeLowercase{\textit{et al.}}: Bare Demo of IEEEtran.cls for Computer Society Journals}

\input{abstract.tex}
\input{sec-1-0-main.tex}

\input{sec-2-0-main.tex}

\input{sec-3-0-main.tex}
\input{sec-4-0-main.tex}
\input{sec-5-0-main.tex}
\input{sec-6-0-ack.tex}
\input{sec-7-0-ref.tex}
\input{sec-8-0-authors.tex}
\end{document}

%% file: abstract.tex
\IEEEtitleabstractindextext{%
\begin{abstract}
Prior studies show that the key to face anti-spoofing lies in the subtle image pattern, termed ``spoof trace", \textit{e.g.}, color distortion, $3$D mask edge, Moiré pattern, and many others. 
Designing a generic face anti-spoofing model to estimate those spoof traces can improve not only the generalization of the spoof detection, but also the interpretability of the model's decision.
Yet, this is a challenging task due to the diversity of spoof types and the lack of ground truth in spoof traces. 
In this work, we design a novel adversarial learning framework to disentangle spoof faces into the spoof traces and the live counterparts. 
Guided by physical properties, the spoof generation is represented as a combination of additive process and inpainting process.
Additive process describes spoofing as spoof material introducing extra patterns (\textit{e.g.}, moire pattern), where the live counterpart can be recovered by removing those patterns.
Inpainting process describes spoofing as spoof material fully covering certain regions, where the live counterpart of those regions has to be ``guessed".
We use 3 additive components and 1 inpainting component to represent traces at different frequency bands.
The disentangled spoof traces can be utilized to synthesize realistic new spoof faces after proper geometric correction, and the synthesized spoof can be used for training and improve the generalization of spoof detection.
Our approach demonstrates superior spoof detection performance on $3$ testing scenarios: known attacks, unknown attacks, and open-set attacks. Meanwhile, it provides a visually-convincing estimation of the spoof traces.
Source code  and pre-trained models will be publicly available upon publication.

\end{abstract}

}
\maketitle
\IEEEdisplaynontitleabstractindextext
\IEEEpeerreviewmaketitle

%% file: sec-1-0-main.tex
\Section{Introduction}
\label{sec:intro}
\IEEEPARstart{I}{n} recent years, the vulnerability of face biometric systems has been widely recognized and increasingly brought attention to the computer vision community.
The attacks to the face biometric systems attempt to deceive the systems
to make wrong identity recognition: either recognize the attackers as a target person (\textit{i.e.}, impersonation), or cover up the original identity (\textit{i.e.}, obfuscation).
\input{figures/figure1}
There are various types of digital and physical attacks, including face morphing~\cite{dale2011video,zakharov2019few,zollhofer2018state}, face adversarial attacks~\cite{deb2019advfaces,goodfellow2014explaining,szegedy2013intriguing}, face manipulation attacks (\textit{e.g.}, deepfake, face swap)~\cite{deepfake,thies2016face2face}, and face spoofing~\cite{bigun2004assuring,frischholz2003avoiding,schuckers2002spoofing}. 
Among the above-mentioned attacks, face spoofing is the only physical attack to deceive the systems, where attackers present faces from spoof mediums, such as photograph, screen, mask and makeup, instead of a live human.
These spoof mediums can be easily manufactured by ordinary people, and hence they pose huge threats to face biometric applications such as mobile face unlock, building access control, and transportation security.
Therefore, face biometric systems need to be secured with face anti-spoofing (FAS) techniques to distinguish the source of the face before performing the face recognition task.
\input{figures/figure2new}

As most face recognition systems are based on a monocular RGB camera, monocular RGB based face anti-spoofing has been studied for over a decade, and one of the most common approaches is based on texture analysis~\cite{boulkenafet2015face,boulkenafet2016face,patel2016secure}.
Researchers noticed that presenting faces from spoof mediums introduces special texture differences, such as color distortions, unnatural specular highlights, Moiré patterns, \textit{etc}.
Those texture differences are inherent within spoof mediums and thus hard to remove or camouflage.
Conventional approaches build a feature extractor plus classifier pipeline, such as LBP+SVM and HOG+SVM~\cite{de2012lbp,komulainen2013context}, and show good performance on several small databases with constraint environments.
In recent years, many works leverage deep learning techniques and show great progress in face anti-spoofing performance~\cite{atoum-depth-fas,liu-auxiliary-fas,deep-tree,shao2019multi,yang2019face}.
Deep learning based methods can be generally grouped into $3$ categories: direct FAS, auxiliary FAS, and generative FAS, as illustrated in Fig.~\ref{fig:2new}.
Early works~\cite{yang2014learn,xu2015learning} build vanilla CNN with binary output to directly predict the spoofness of an input face (Fig.\ref{fig:2new}a). 
Methods~\cite{liu-auxiliary-fas,yang2019face} propose to learn an intermediate representation, \textit{e.g.}, depth, rPPG, reflection, instead of binary classes, which can lead to better generalization and performance (Fig.\ref{fig:2new}b).
~\cite{jourabloo-face-despoofing, joel2020cvpr, feng2020learning} additionally attempt to generate the visual patterns existing in the spoof samples (Fig.\ref{fig:2new}c), providing a more intuitive interpretation of the sample's spoofness.

Despite the success, there are still at least three unsolved problems in the topic of deep learning-based face anti-spoofing. 
First, most prior works are designed to tackle {\it limited spoof types}, either print/replay or $3$D mask solely, while a real-world anti-spoofing system may encounter a wide variety of spoof types including print, replay, various $3$D masks, facial makeup, and even unseen attack types. Therefore, to better reflect real-world performance, we need a benchmark to evaluate face anti-spoofing under known attacks, unknown attacks, and their combination (termed {\it open-set} setting).
Second, many approaches formulate face anti-spoofing as a classification/regression problem, with a single score as the output. 
Although auxiliary FAS and generative FAS attempt to offer some extent of interpretation by fixation, saliency, or noise analysis, there is little understanding on what the exact differences are between live and spoof, and what patterns the classifier's decision is based upon. 
A better interpretation can be estimating the exact patterns differentiating a spoof face and its live counterpart, termed \textbf{spoof trace}. 
Thirdly, compared with other face analysis tasks such as recognition or alignment, the data for face anti-spoofing has several limitations. Most FAS databases are captured in the constraint indoor environment, which has limited intra-subject variation and environment variation. For some special spoof types such as makeup and customized silicone mask, they
require highly skilled experts to apply or create, with high cost, which results in very limited samples (\textit{i.e.}, long-tail data). Thus, how to learn from data with limited variations or samples is a challenge for FAS.


In this work, we aim to design a face anti-spoofing model that is applicable to a wide variety of spoof types, termed \textbf{generic face anti-spoofing}. We equip this model with the ability to explicitly disentangle the spoof traces from the input faces. 
Some examples of spoof trace disentanglement are shown in Fig.~\ref{fig:1}. 
This is a challenging objective due to the diversity of spoof traces and the lack of ground truth during model learning. However, we believe that fulfilling this objective can bring  several benefits:
\begin{enumerate}
    \item Binary classification for face anti-spoofing would harvest any cue that helps classification, which might include spoof-irrelevant cues such as lighting, and thus hinder generalization. In contrast, spoof trace disentanglement explicitly tackles the most fundamental cue in spoofing, upon which the classification can be more grounded and witness better generalization.
    \item With the trend of pursuing explainable AI~\cite{darpa-xai,arrieta2020explainable}, it is desirable for the face anti-spoofing model to generate the spoof patterns that support its binary decision,
    since spoof trace serves as a good visual explanation of the model's decision. Certain properties (\textit{e.g.}, severity, methodology) of spoof attacks might potentially be revealed from the traces.
    \item Disentangled spoof traces can enable the synthesis of realistic spoof samples, which addresses the issue of limited training data for the minority spoof types, such as special $3$D masks and makeup. 
\end{enumerate}

As shown in Fig.~\ref{fig:2new}d, we propose a Physics-guided Spoof Trace Disentanglement (PhySTD) to explore the spoof traces for generic face anti-spoofing.
To model all types of spoofs, we formulate the spoof trace disentanglement as a combination of {\it additive} process and {\it inpainting} process. 
Additive process describes spoofing as spoof material introducing extra patterns (\textit{e.g.}, moire pattern), where the live counterpart can be recovered by removing those patterns.
Inpainting process describes spoofing as spoof material fully covering certain regions of the original face, where the live counterpart of those regions has to be ``guessed"~\cite{liu2015comparison,bertalmio2000image}.
We further decompose the spoof traces into frequency-dependent components, so that traces with different frequency properties can be equally handled.
For the network architecture, we extend a backbone network for auxiliary FAS with a decoder to perform the disentanglement. With no ground truth of spoof traces, we adopt an overall GAN-based training strategy.
The generator takes an input face, estimates its spoofness, and disentangles the spoof trace.
After obtaining the spoof trace, we can reconstruct the live counterpart from the spoof and synthesize new spoof from the live.
The synthesized samples are then sent to multiple discriminators with real samples for adversarial training.
The synthesized spoof samples are further utilized to train the generator in a fully supervised fashion, thanks to disentangled spoof traces as ground truth for the synthesized samples.
To correct possible geometric discrepancy during spoof synthesis, we propose a novel $3$D warping layer to deform spoof traces toward the target live face.

A preliminary version of this work was published in the
Proceedings European Conference on Computer Vision (ECCV)  2020~\cite{liu2020on}. 
We extend the work from three aspects. 
$1$) Guided by the physics of how a spoof is generated, we introduce a spoof generation function (SGF) to model the spoof trace disentanglement as a combination of additive and inpainting processes. SGF has a better and more natural modeling of generic spoof attacks, such as paper glass.
$2$) Previous trace components $\{\mathbf{S},\mathbf{B},\mathbf{C},\mathbf{T}\}$ are not supervised hierarchically so that there exists semantic ambiguity. In this work, we introduce several hierarchical designs in the GAN framework to remedy such ambiguity.
$3$) We propose an open-set testing scenario to further evaluate the real-world performance for face anti-spoofing models. Both known and unknown attacks are included in the open-set testing. We perform a side-by-side comparison between the proposed approach and the state-of-the-art (SOTA) face anti-spoofing solutions on multiple datasets and protocols.

In summary, the main contributions of this work are as follows:
\begin{itemize}
\item[$\bullet$] We for the first time study spoof trace for {\it generic} face anti-spoofing, where a wide variety of spoof types are tackled with one unified framework; 
\item[$\bullet$] We propose a novel physics-guided model to disentangle spoof traces, and utilize the spoof traces to synthesize new data samples for enhanced training; 
\item[$\bullet$] We propose novel protocols for a generic open-set face anti-spoofing;
\item[$\bullet$] We achieve SOTA anti-spoofing performance and provide convincing visualization for a wide variety of spoof types. 
\end{itemize}

%% file: figures/figure1.tex
\begin{figure}[t]
    \centering
    \resizebox{1\linewidth}{!}{\includegraphics{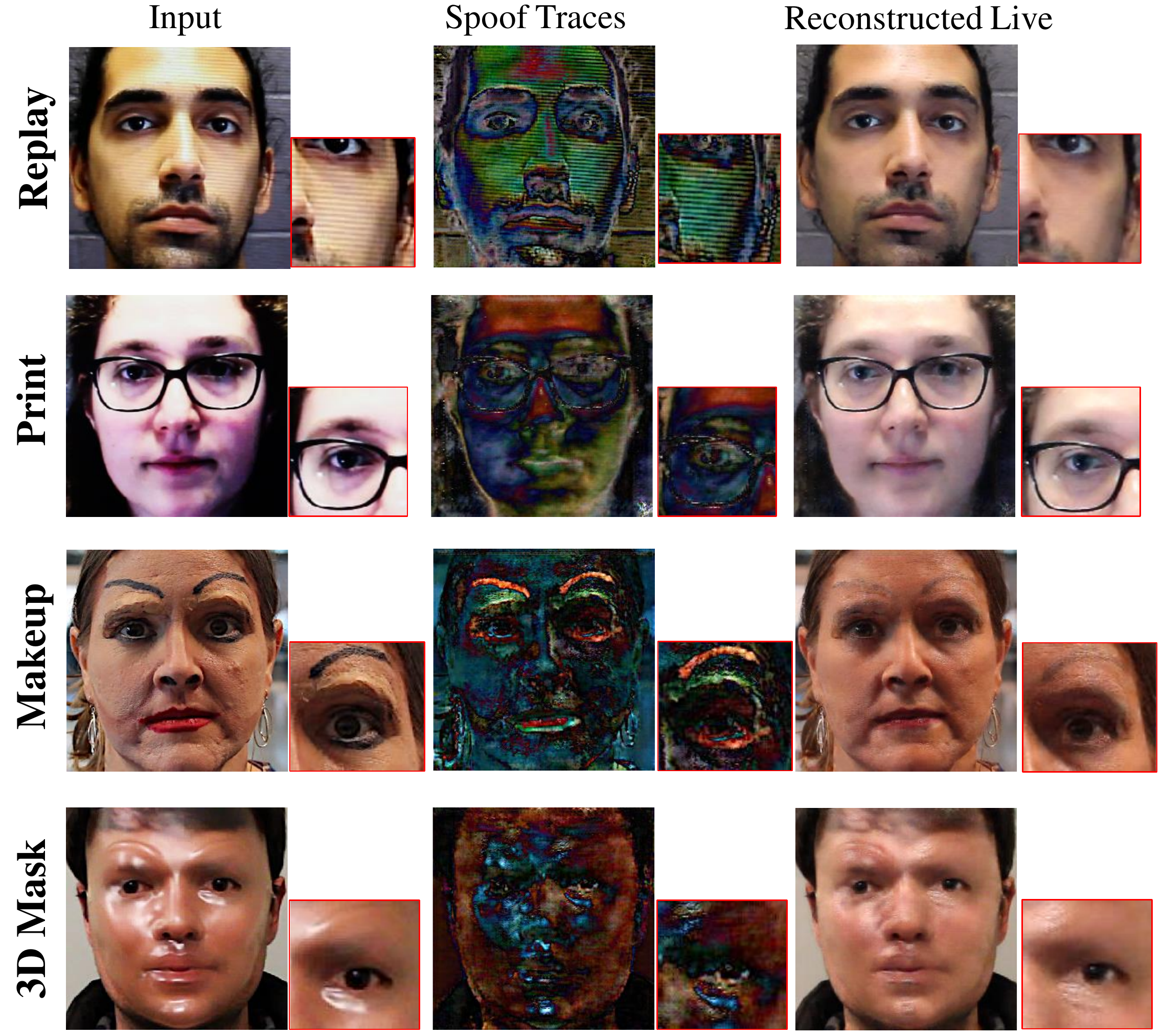}}
    \caption{\small The proposed approach can detect spoof faces, disentangle the spoof traces, and reconstruct the live counterparts. It can be applied to diverse spoof types and estimate distinct traces ({\it e.g.}, Moiré pattern in replay attack, artificial eyebrow and wax in makeup attack, color distortion in print attack, and specular highlights in $3$D mask attack). Zoom in for details.}
    \label{fig:1}
\end{figure}

%% file: figures/figure2new.tex
\begin{figure*}[t]
    \centering
    \resizebox{1\linewidth}{!}{\includegraphics{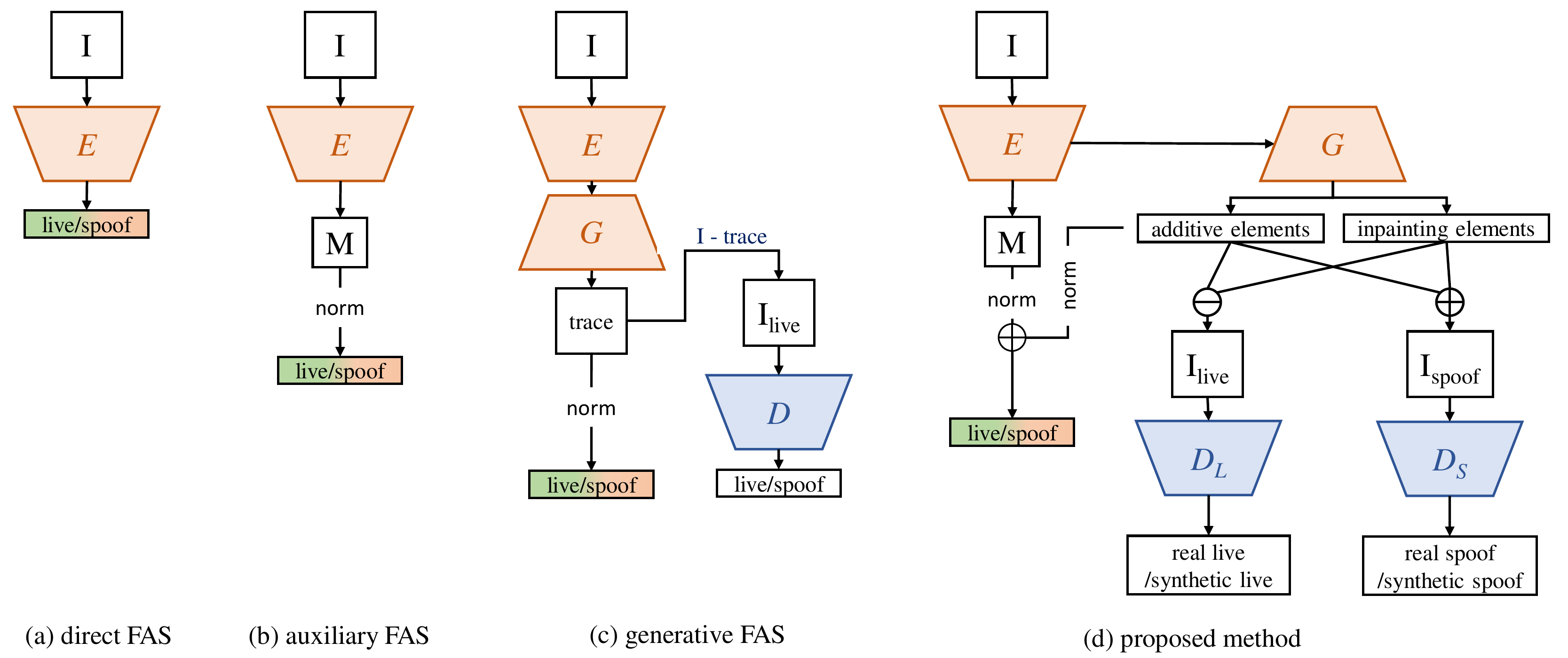}}
    \caption{\small The comparison of different deep-learning based face anti-spoofing. (a) direct FAS only provides a binary decision of spoofness; (b) auxiliary FAS can provide simple interpretation of spoofness. $\mathbf{M}$ denotes the auxiliary task, such as depth map estimation; (c) generative FAS can provide more intuitive interpretation of spoofness, but only for a limited number of spoof attacks; (d) the proposed method can provide spoof trace estimation for generic face spoof attacks. }
    \label{fig:2new}
\end{figure*}

%% file: sec-2-0-main.tex
\Section{Related Work}
\label{sec:prior}
\Paragraph{Face Anti-Spoofing}
Face anti-spoofing has been studied for more than a decade and its development can be roughly divided into three stages.
In the early years, researchers leverage spontaneous human movement, such as eye blinking and head motion, to detect simple print photograph or static replay attacks~\cite{kollreider2007real,pan2007eyeblink}. 
However, when facing counter attacks, such as print face with eye region cut, and replaying a face video, those methods would  fail.
In the second stage, researchers pay more attention to texture differences between live and spoof, which are inherent to spoof mediums.
Researchers mainly extract hand-crafted features from the faces, \textit{e.g.}, LBP~\cite{boulkenafet2015face,de2012lbp,de2013can,maatta2011face}, HoG ~\cite{komulainen2013context,yang2013face}, SIFT~\cite{patel2016secure} and SURF~\cite{boulkenafet2016face}, and train a classifier to split the live \textit{vs.}~spoof, \textit{e.g.}, SVM and LDA.

Recently, face anti-spoofing solutions equipped with deep learning techniques have demonstrated significant improvements over the conventional methods. 
Methods in~\cite{feng2016integration,li2016original,patel2016cross,yang2014learn} train a deep neural network to learn a binary classification between live and spoof.
In~\cite{atoum-depth-fas,liu-auxiliary-fas,deep-tree,shao2019multi,yang2019face}, additional supervisions, such as face depth map and rPPG signal, are utilized to help the network to learn more generalizable features.
As the latest approaches achieving saturated performance on several benchmarks, researchers start to explore more challenging cases, such as few-shot/zero-shot face anti-spoofing~\cite{deep-tree,qin2019learning,zhao2019meta} and domain adaptation in face anti-spoofing~\cite{shao2019multi,shao2019regularized}.

In this work, we aim to solve an interesting yet very challenging problem: disentangling and visualizing the spoof traces from an input face. 
A related work~\cite{jourabloo-face-despoofing} also adopts GAN seeking to estimate the spoof traces. However, they formulate the traces as low-intensity noises, which is limited to print and replay attacks only and cannot provide convincing visual results. 
In contrast, we explore spoof traces for a much wider range of spoof attacks, visualize them with novel disentanglement, and also evaluate the proposed method on the challenging cases, \textit{e.g.}, zero-shot face anti-spoofing.

\Paragraph{Disentanglement Learning} Disentanglement learning is often adopted to better represent complex data and features.
DR-GAN~\cite{disentangled-representation-learning-gan-for-pose-invariant-face-recognition} disentangles a face into identity and pose vectors for pose-invariant face recognition and view synthesis.
Similarly in gait recognition, ~\cite{gait-recognition-via-disentangled-representation-learning} disentangles the representations of appearance, canonical, and pose features from an input gait video.
$3$D reconstruction works~\cite{disentangling-features-in-3d-face-shapes-for-joint-face-reconstruction-and-recognition, on-learning-3d-face-morphable-model-from-in-the-wild-images} also disentangle the representation of a $3$D face into identity, expressions, poses, albedo, and illuminations.
For image synthesis, \cite{esser2018variational} disentangles an image into appearance and shape with U-Net and Variational Auto Encoder (VAE).

Different from~\cite{disentangling-features-in-3d-face-shapes-for-joint-face-reconstruction-and-recognition,disentangled-representation-learning-gan-for-pose-invariant-face-recognition,gait-recognition-via-disentangled-representation-learning}, we intend to disentangle features that have different scales and contain geometric information. We leverage the multiple outputs to represent features at different scales, and adopt multiple-scale discriminators to properly learn them. Moreover, we propose a novel warping layer to tackle the geometric discrepancy during the disentanglement and reconstruction.

\Paragraph{Image Trace Modeling}
Image traces are certain signals existing in the image that can reveal information about the capturing camera, imaging setting, environment, and so on. 
Those signals often have much lower energy compared to the image content, which needs proper modeling to explore them.
\cite{abdelhamed2018high,thai2013camera,thai2016camera} observe the difference of image noises, and use them to recognize the capture cameras.
From the frequency domain, \cite{joel2020cvpr} shows the image noises from different cameras obey different noise distributions.
Such techniques are applied to the field of image forensics, and later \cite{wang2017counter,chen2020camera} propose methods to remove such traces for image anti-forensics.

Recently, image trace modeling is widely used in image forgery detection and image adversarial attack detection~\cite{wu2019mantra,dang2020detection}. In this work, we attempt to explore the traces of spoof face presentation. 
Due to different spoof mediums, spoof traces show large variations in content, intensity, and frequency distribution. We propose to disentangle the traces as additive traces and inpainting trace. And for additive traces, we further decompose them based on different frequency bands.

%% file: sec-3-0-main.tex
\input{figures/figure2}

\Section{Physics-based Spoof Trace Disentanglement}\label{sec:method}
\SubSection{Problem Formulation}
\label{sec:3-1}
Let the domain of live faces be denoted as $\mathcal{L}\! \subset \! \mathbb{R}^{N\!\times \!N \!\times \!3}$ and spoof faces as $\mathcal{S}\! \subset \!\mathbb{R}^{N\!\times \!N\! \times \!3}$, where $N$ is the image size.
We intend to obtain not only the correct prediction (live \textit{vs.} spoof) of the input face, but also a convincing estimation of the spoof trace and live face reconstruction.
To represent the spoof trace, our preliminary version assumes an additive relation between live and spoof, and uses $4$ trace components $\{\mathbf{S},\mathbf{B},\mathbf{C},\mathbf{T}\}$ at different frequency bands as:
\begin{equation}
\label{eq:recon_old}
\mathbf{I}_{\textit{spoof}} = (1+\lfloor\mathbf{S}\rfloor_{n_1})\mathbf{I}_{\textit{live}} + \lfloor\mathbf{B}\rfloor_{n_1} + \lfloor\mathbf{C}\rfloor_{n_2} + \mathbf{T},
\end{equation}
where $\mathbf{S},\mathbf{B}$ represent low-frequency traces, $\mathbf{C}$ represents mid-frequency ones, and $\mathbf{T}$ represents high-frequency ones.
$\lfloor\cdot\rfloor$ is the low bandpass filtering operation, and in practice, we achieve this by downsampling the original image and upsampling it back.  
In the previous setting, $n_1\!=\!1$ and $n_2\!=\!64$.
Compared to the simple representation with only a single component~\cite{jourabloo-face-despoofing}, this multi-scale representation of $\{\mathbf{S},\mathbf{B},\mathbf{C},\mathbf{T}\}$ can largely improve disentanglement quality and suppress undesired artifacts due to its coarse-to-fine process.
The model is designed to provide a valid estimation of spoof traces $\{\mathbf{S},\mathbf{B},\mathbf{C},\mathbf{T}\}$ without respective ground truth. 
Our preliminary version~\cite{liu2020on} aims to find a minimum intensity change that transfers an input face to the live domain: 
\begin{equation}
\label{eq:op_old}
\argmin_{\hat{\mathbf{I}}} \| \mathbf{I} - \hat{\mathbf{I}}\|_F \; s.t. \; \mathbf{I} \in (\mathcal{S}\cup\mathcal{L}) \; \text{and} \; \hat{\mathbf{I}} \in \mathcal{L},
\end{equation}
where $\mathbf{I}$ is the source face, $\hat{\mathbf{I}}$ is the target face to be optimized, and $\mathbf{I}-\hat{\mathbf{I}}$ is defined as the spoof trace.
When the source face is live $\mathbf{I}_{\text{live}}$, $\mathbf{I}-\hat{\mathbf{I}}$ should be $0$ as $\mathbf{I}$ is already in $\mathcal{L}$. When the source face is spoof $\mathbf{I}_{\text{spoof}}$, $\mathbf{I}-\hat{\mathbf{I}}$ should be regularized to prevent unnecessary changes such as identity shift.

Despite the effectiveness of this representation, there are still two drawbacks:
First, the spoof trace disentanglement is mainly formulated as an additive processing. The optimization of Eqn.~\ref{eq:op_old} limits the trace intensity, and the reconstruction for spoof regions with large appearance divergence might be sub-optimal, such as spoof glasses or mask. For those spoof regions, the physical relationship between the live and the spoof is better described as replacement rather than addition; 
Second, while  our preliminary version representing the traces with hierarchical components, these components are learned with losses on their summation. 
Without careful supervision, the learned components can be ambiguous in their semantic meanings, {\it e.g.}, the high-frequency component may include low-frequency information. 

To address the first drawback, we introduce a spoof generation function (SGF) as an additive process followed by an inpainting process:
\begin{equation}
\label{eq:spoof_generation}
\mathbf{I}_{\textit{spoof}} = (1-\mathbf{P})(\mathbf{I}_{\textit{live}} +\mathbf{T}_A) + \mathbf{P}\cdot\mathbf{T}_P,
\end{equation}
where $\mathbf{T}_A\!\in \mathbb{R}^{N\!\times \!N\!\times \!3}$ indicates the traces from additive process, $\mathbf{T}_P$ indicates the traces from inpainting process, and $\mathbf{P} \in \mathbb{R}^{N\!\times \!N\!\times \!1}$ denotes the inpainting region.
Given a spoof face, one may reconstruct the live counterpart by inversing Eqn.~\ref{eq:spoof_generation}:
\begin{equation}
\label{eq:recon} 
\hat{\mathbf{I}}_{\textit{live}} =(1-\mathbf{P})(\mathbf{I}_{\textit{spoof}} - \mathbf{T}_\text{A}) + \mathbf{P}\cdot(\hat{\mathbf{I}}_{\textit{live}} + \mathbf{I}_{\textit{spoof}} - \mathbf{T}_P),
\end{equation}
As the inpainting physically replaces content, the spoof trace $\mathbf{T}_P$ in the inpainting region $\mathbf{P}$ is identical to the spoof image $\mathbf{I}_{\textit{spoof}}$ in the same region, and thus both cancel out in the second term of Eqn.~\ref{eq:recon}. 
We further rename the $\hat{\mathbf{I}}_{\textit{live}}$ in the second term as $\mathbf{I}_P$ to indicate the inpainting content within the inpainting region that should be estimated from the model. 
Therefore, the reconstruction of the live image becomes:
\begin{equation}
\label{eq:recon2} 
\hat{\mathbf{I}}_{\textit{live}} =(1-\mathbf{P})(\mathbf{I}_{\textit{spoof}} - \mathbf{T}_\text{A}) + \mathbf{P}\cdot\mathbf{I}_P,
\end{equation}
where $\mathbf{T}_A \!=\! \lfloor\mathbf{B}\rfloor_{n_1} \!+\! \lfloor\mathbf{C}\rfloor_{n_2}\!+\! \mathbf{T}$ denotes the additive trace represented by three hierarchical components.
$n_1$ and $n_2$ are set to be $32$ and $128$ respectively.
With a larger $n_1$, the effect of component $\mathbf{S}$ in the preliminary version can be incorporated into $\mathbf{B}$, and hence we remove $\mathbf{S}$ for simplicity.
Besides the additive traces, the model is further required to estimate the inpainting region $\mathbf{P}$ and inpainting live content $\mathbf{I}_P$. $\mathbf{I}_P$ is estimated based on the rest of the live facial region without intensity constraint.
We use a function $G(\cdot)$ to represent the reconstruction process of Eqn.~\ref{eq:recon2}.
Accordingly, the optimization of Eqn.~\ref{eq:op_old} is re-formulated by replacing $\hat{\mathbf{I}}$ with Eqn.~\ref{eq:recon2} as:
\begin{equation}
\begin{gathered}\label{eq:op}
\argmin_{\mathbf{T}_A,\mathbf{P},\mathbf{I}_P} \| \mathbf{I}-(1-\mathbf{P})(\mathbf{I} -\mathbf{T}_A) - \mathbf{P}\cdot\mathbf{I}_P\|_F \;  \\
\rightarrow\argmin_{\mathbf{T}_A,\mathbf{P},\mathbf{I}_P} \|(1-\mathbf{P})\mathbf{T}_A\|_F + \|\mathbf{P}\cdot(\mathbf{I}-\mathbf{I}_P)\|_F. \;  \\
    \end{gathered}
\end{equation}
As we do not wish to impose any intensity constraint on $\mathbf{I}_P$, the final objective is formulated as:
\begin{equation}
    \begin{gathered}
\label{eq:op_final}
\argmin_{\mathbf{T}_A,\mathbf{P}} \| (1\!-\!\mathbf{P})\mathbf{T}_A\|_F+\lambda\| \mathbf{P}\|_F \;
s.t. \; \mathbf{I} \! \in \! \mathcal{S}\cup\mathcal{L}, \hat{\mathbf{I}} \in \mathcal{L},
    \end{gathered}
\end{equation}
where $\lambda$ is a weight to balance two terms.
In addition, based on Eqn.~\ref{eq:spoof_generation}, we can define another function $G^-(\cdot)$ to synthesize new spoof faces, by transferring the spoof traces from $\mathbf{I}^i$ to $\mathbf{I}^j$:
\begin{equation}
\label{eq:syn}
\hat{\mathbf{I}}_{\textit{spoof}}^{i\rightarrow j} = G^-(\mathbf{I}^j|\mathbf{I}^i) = (1-\mathbf{P}^i)(\mathbf{I}^j+ \mathbf{T}_\text{A}^i) + \mathbf{P}^i\cdot\mathbf{I}^i.
\end{equation}
Note that $\mathbf{T}_P$ in Eqn.~\ref{eq:spoof_generation} has been replaced with $\mathbf{I}^i$ since the spoof image $\mathbf{I}^i$ contains the spoof trace for the inpainting region.

\input{figures/figure3}

Estimating $\{\mathbf{T}_\text{A},\mathbf{P},\mathbf{I}_P\}$ from an input face $\mathbf{I}$ is termed as \textbf{spoof trace disentanglement}.
Given that no ground truth of traces is available, this disentanglement can be achieved via generative adversarial based training.
As shown in Fig.~\ref{fig:2}, the proposed Physics-guided Spoof Trace Disentanglement (PhySTD) consists of a generator and discriminator. 
Given an input image, the generator is designed to predict the spoofness (represented by the pseudo depth map) as well as estimate the additive traces $\{\mathbf{B}, \mathbf{C}, \mathbf{T}\}$ and the inpainting components $\{\mathbf{P},\mathbf{I}_P\}$. 
With the traces, we can apply function $G(\cdot)$ to reconstruct the live counterpart and function $G^-(\cdot)$ to synthesize new spoof faces. 
We adopt a set of discriminators at multiple image resolutions to distinguish the real faces $\{\mathbf{I}_{\textit{live}},\mathbf{I}_{\textit{spoof}}\}$ with the synthetic faces $\{\hat{\mathbf{I}}_{\textit{live}},\hat{\mathbf{I}}_{\textit{spoof}}\}$.
To remedy the semantic ambiguity during $\{\mathbf{B}, \mathbf{C}, \mathbf{T}\}$ learning, three trace component combinations, $\{\mathbf{B}\}$, $\{\mathbf{B}, \mathbf{C}\}$, and $\{\mathbf{B}, \mathbf{C}, \mathbf{T}\}$, will contribute to the synthesis of live reconstruction at one particular resolution, which is then supervised by a respective discriminator (details in Sec.\ref{sec:recon}).
To learn a proper inpainting region $\mathbf{P}$, we leverage both the prior knowledge and the information from the additive traces.

In the rest of this section, we present the details of the generator, the discriminators, the details of face reconstruction and synthesis, and the losses and training steps used in PhySTD.

\input{sec-3-1-gen}
\input{sec-3-2-recon}
\input{sec-3-3-dis}
\input{sec-3-4-loss}


%% file: figures/figure2.tex
\begin{figure*}[t!]
    \centering
    \resizebox{1\linewidth}{!}{\includegraphics{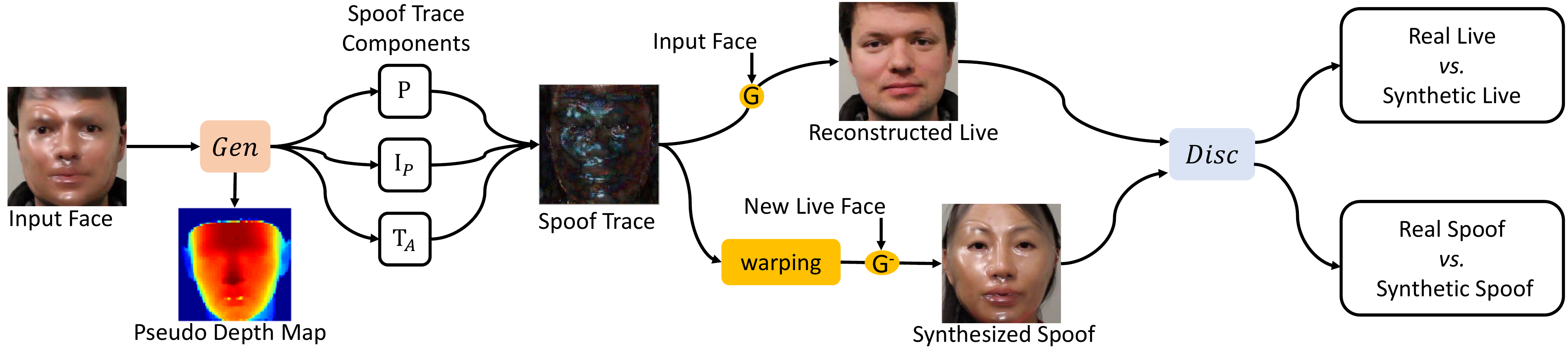}}
    \caption{\small Overview of the proposed Physics-guided Spoof Trace Disentanglement (PhySTD).}
    \label{fig:2}
\end{figure*}

%% file: figures/figure3.tex
\begin{figure*}[t!]
    \centering
    \resizebox{\linewidth}{!}{\includegraphics{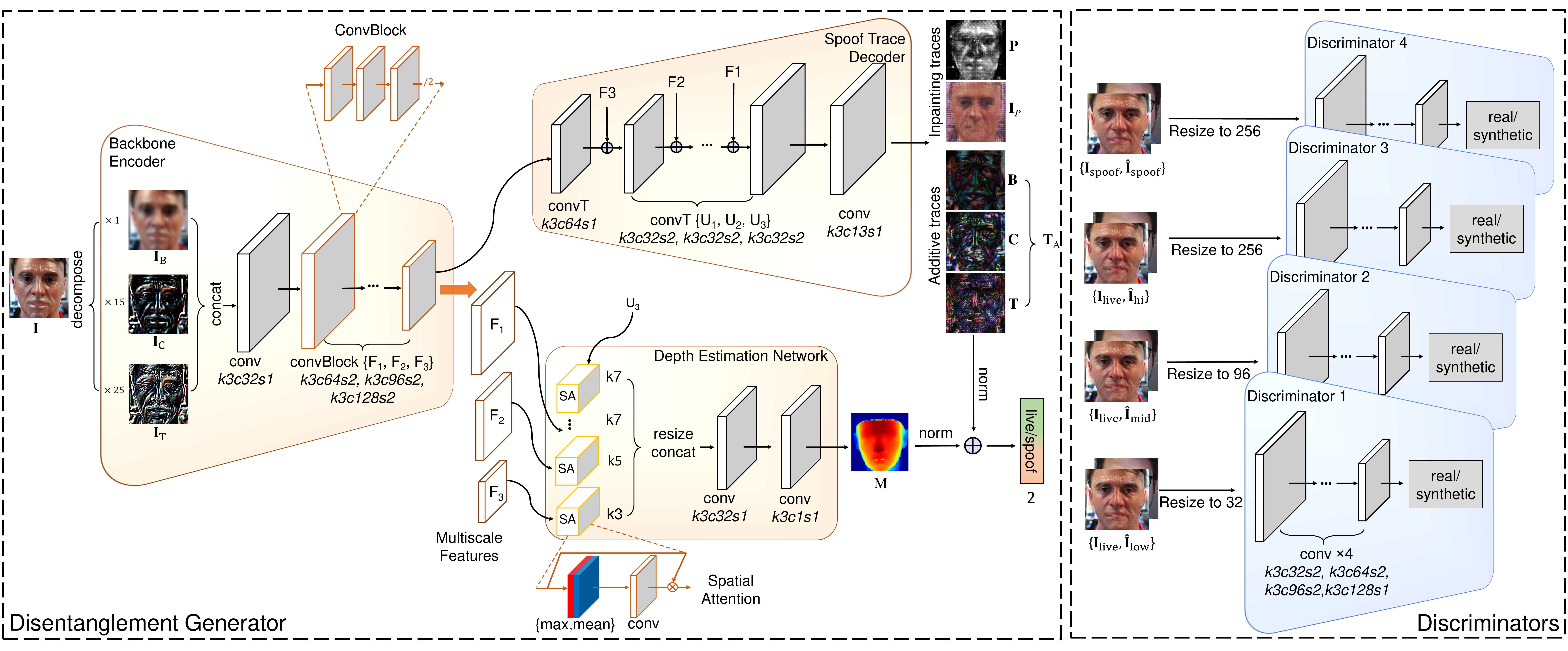}}
    \caption{\small The proposed PhySTD network architecture. Except the last layer, each conv and transposed conv is concatenated with a batch normalizstion layer and a leaky ReLU layer. \textit{‘k3c64s2’} indicates the kernel size of $3\times3$, the convolution channel of $64$ and the stride of $2$. 
    }
    \label{fig:3}
\end{figure*}

%% file: sec-3-1-gen.tex
\SubSection{Disentanglement Generator}
As shown in Fig.~\ref{fig:3}, the disentanglement generator consists of a backbone encoder, a spoof trace decoder and a depth estimation network. The backbone encoder aims to extract multi-scale features, the depth estimation network leverages the features to estimate the facial depth map, and a spoof trace decoder to estimate the additive trace components $\{\mathbf{B},\mathbf{C},\mathbf{T}\}$ and the inpainting components $\{\mathbf{P},\mathbf{I}_P\}$. 
The depth map and the spoof traces will be used to compute the final spoofness score.
\input{figures/figure3-2}

\Paragraph{Backbone encoder}
Backbone encoder extracts features from the input images for both depth map estimation and spoof trace disentanglement.
As shown in our preliminary work~\cite{liu2020on}, the spoof traces consists of components from different frequency bands: low-frequency traces includes color distortion, mid-frequency traces includes makeup strikes, and high-frequency traces includes Moiré patterns and mask edges. However, a vanilla CNN model might overlook high-frequency traces since the energy of high-frequency traces is often much weaker than that of low-frequency traces.
In order to encourage the network to equally regard traces with different physical properties, we explicitly decompose the image into three elements $\{\mathbf{I}_\mathbf{B},\mathbf{I}_\mathbf{C},\mathbf{I}_\mathbf{T}\}$ as:
\begin{equation}
\label{eq:deco}
\begin{split}
    \mathbf{I}_\mathbf{B}=&\lfloor\mathbf{I}\rfloor_{n_1},\\
    \mathbf{I}_\mathbf{C}=&\lfloor\mathbf{I}\rfloor_{n_2}-\lfloor\mathbf{I}\rfloor_{n_1},\\
    \mathbf{I}_\mathbf{T}=& \mathbf{I} - \lfloor\mathbf{I}\rfloor_{n2}, \\
\end{split}
\end{equation}
where $n_1\!=\!32$, $n_2\!=\!128$ and the image size $N\!=\!256$. In addition, we amplify the value in $\mathbf{I}_\mathbf{C}, \mathbf{I}_\mathbf{T}$ by two constants $15$ and $25$, and then feed the concatenation of three elements to the backbone network.
Fig.~\ref{fig:3-2} provides the visualization of image decomposition. We observe that the traces that are less distinct in the original images become more highlighted in the $\mathbf{I}_\mathbf{T}$ component: $3$D mask and replay attack bring unique patterns different with the live face pattern, while print attack is lacking of necessary high frequency details. 
Semantically, $\mathbf{I}_\mathbf{B},\mathbf{I}_\mathbf{C},\mathbf{I}_\mathbf{T}$ share the same frequency domains with $\mathbf{B},\mathbf{C},\mathbf{T}$ respectively, and thus the decomposition potentially eases the learning of $\mathbf{B},\mathbf{C},\mathbf{T}$. 

After that, the encoder progressively downsamples the decomposed image components $3$ times to obtain features  $\mathbf{F}_1\! \in \!\mathbb{R}^{128\!\times \!128\!\times \!64}$, $\mathbf{F}_2\! \in \!\mathbb{R}^{64\!\times \!64\!\times \!96}$, $\mathbf{F}_3\! \in \!\mathbb{R}^{32\!\times \!32\!\times \!128}$ via conv layers.

\Paragraph{Spoof trace decoder}
The decoder upsamples the feature $\mathbf{F}_3$ with transpose conv layers back to the input face size $256$.
The last layer outputs both additive traces $\{\mathbf{B},\mathbf{C},\mathbf{T}\}$ and inpainting components $\{\mathbf{P},\mathbf{I}_P\}$.
Similar to U-Net~\cite{ronneberger2015u}, we apply the short-cut connection between the backbone encoder and decoder to bypass the multiple scale details for a high-quality trace estimation.

\Paragraph{Depth estimation network}
We still recognize the importance of the discriminative supervision used in auxiliary FAS, and thus introduce a depth estimation network to perform the pseudo-depth estimation for face anti-spoofing, as proposed in~\cite{liu-auxiliary-fas}.
The depth estimation network takes the concatenated features of $\mathbf{F}_1$, $\mathbf{F}_2$, $\mathbf{F}_3$ from the backbone encoder and $\mathbf{U}_3$ from the decoder as input. The features are put through a spatial attention mechanism from~\cite{cdcn} and resize to the same size of $K=32$. It outputs a face depth map $\mathbf{M}\!\in\! \mathbb{R}^{32\!\times \!32}$, where the depth values are normalized within $[0,1]$. 
Regarding the number of parameters, both spoof trace decoder and depth estimation network are light weighed, while the backbone network is much heavier.
With more network layers being shared to tackle both depth estimation and spoof trace disentanglement, the knowledge learnt from spoof trace disentanglement can be better shared with depth estimation task, which can lead to a better anti-spoofing performance.

\Paragraph{Final scoring}
In the testing phase, we use the norm of the depth map and the intensity of spoof traces for real vs.~spoof classification:
\eqnvspace
\begin{equation}
\label{eq:score}
\text{score} = \frac{1}{2K^2}\|\mathbf{M}\|_1 \!+ \!
			   \frac{\alpha_0}{2N^2}(\|\mathbf{B}\|_1\!+\!\|\mathbf{C}\|_1\!+\!\|\mathbf{T}\|_1\!+\!\|\mathbf{P}\|_1),
\end{equation}
where $\alpha_0$ is the weight for the spoof trace.


%% file: figures/figure3-2.tex
\begin{figure}[t!]
    \centering
    \resizebox{0.95\linewidth}{!}{\includegraphics{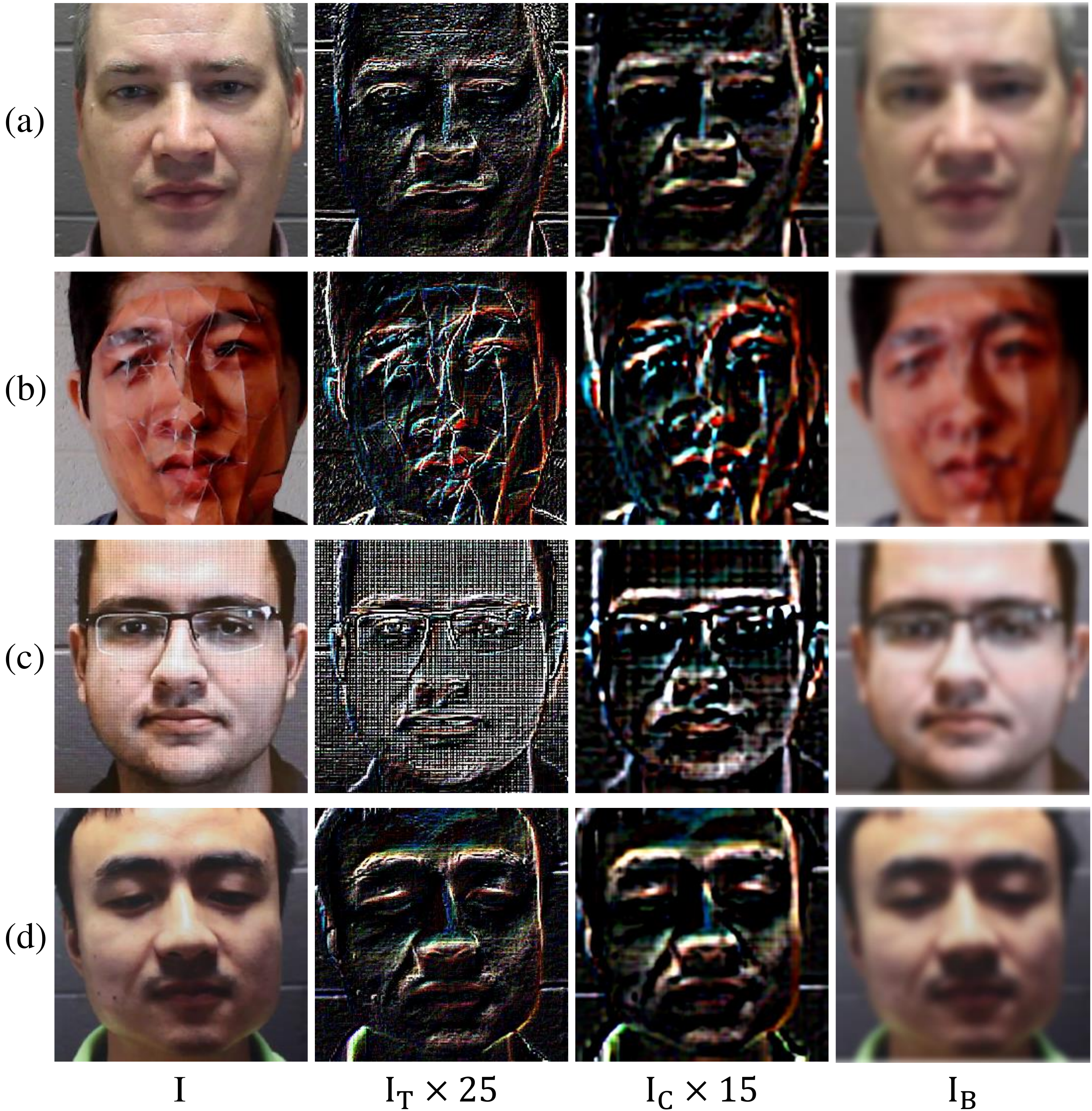}}
    \caption{\small The visualization of image decomposition for different input faces: (a) live face (b) $3$D mask attack (c) replay attack (d) print attack.} 
    \label{fig:3-2}
\end{figure}

%% file: sec-3-2-recon.tex
\SubSection{Reconstruction and Synthesis}
\label{sec:recon}
There are multiple options to use the disentangled spoof traces: $1$)~live reconstruction, $2$)~spoof synthesis, and $3$)~``harder" sample synthesis, which will be described below respectively.

\Paragraph{Live reconstruction:}
Based on Eqn.~\ref{eq:recon2}, we propose a hierarchical reconstruction of the live face counterpart from the input images. 
To reconstruct faces at a certain resolution, each additive trace is included only if its frequency domain is lower than the target resolution. 
We apply $\{\textit{hi},\textit{mid},\textit{low}\}$ three resolution settings as:
\eqnvspace
\begin{equation}
\label{eq:recon_deco}
\begin{split}
    \hat{\mathbf{I}}_{\textit{hi}}=&(1-\mathbf{P})(\mathbf{I} - \lfloor\mathbf{B}\rfloor_{n_1} - \lfloor\mathbf{C}\rfloor_{n_2} - \mathbf{T}) + \mathbf{P}\cdot\mathbf{I}_P,\\
    \hat{\mathbf{I}}_{\textit{mid}}=&(1-\mathbf{P})(\lfloor\mathbf{I}\rfloor_{n_2} - \lfloor\mathbf{B}\rfloor_{n_1} - \lfloor\mathbf{C}\rfloor_{n_2}) + \mathbf{P}\cdot\mathbf{I}_P,\\
    \hat{\mathbf{I}}_{\textit{low}}=&(1-\mathbf{P})(\lfloor\mathbf{I}\rfloor_{n_1} - \lfloor\mathbf{B}\rfloor_{n_1}) + \mathbf{P}\cdot\mathbf{I}_P.\\
\end{split}
\end{equation}

\Paragraph{Spoof synthesis:}
Based on Eqn.~\ref{eq:syn}, we can obtain a new spoof face via applying the spoof traces disentangled from a spoof face $\mathbf{I}_i$ to a live face $\mathbf{I}_j$. 
However, spoof traces may contain face-dependent content associated with the original spoof subject.
Directly applying them to a new face with different shapes or poses may result in mis-alignment and strong visual implausibility. 
Therefore, the spoof trace should go through a geometry correction before performing this synthesis. 
We propose an online $3$D warping layer and will introduce it in the following subsection.

\Paragraph{``Harder" sample synthesis:}
The disentangled spoof traces can not only reconstruct live and synthesize new spoof, but also synthesize ``harder" spoof samples by removing or amplifying part of the spoof traces.
We can tune one or some of the trace elements $\{\mathbf{B},\mathbf{C},\mathbf{T},\mathbf{P}\}$ to make the spoof sample to become ``less spoofed", which is thus closer to a live face since the spoof traces are weakened.
Such spoof data can be regarded as \textit{harder} samples and may benefit the generalization of the disentanglement generator.
For instance, while removing the low frequency element $\mathbf{B}$ from a replay spoof trace, the generator may be forced to rely on other elements such as high-level texture patterns.
To synthesize the ``harder" sample $\hat{\mathbf{I}}_{\textit{hard}}$, we follow Eqn.~\ref{eq:syn} with two minor changes: 1) generate $3$ random weights between $[0,1]$ and  multiple each with one component of $\{\mathbf{B}, \mathbf{C}, \mathbf{T}\}$; 2) randomly remove the inpainting process (\textit{i.e.}, set $\mathbf{P}=0$) with a probability of $0.5$. 
Compared with other methods, such as brightness and contrast change~\cite{presentation-attack-detection-for-face-in-mobile-phones}, reflection and blurriness effect~\cite{yang2019face}, or $3$D distortion~\cite{guo2019improving}, our approach can introduce more realistic and effective data samples, as shown in Sec.~\ref{sec:4}.

\SubSubSection{Online $3$D Warping Layer}
We propose an online $3$D warping layer to correct the shape discrepancy.
To obtain the warping, previous methods in~\cite{chang-asymmetric-style-transfer,liu-auxiliary-fas} use offline face swapping and pre-computed dense offset respectively, where both methods are non-differentiable as well as memory intensive. 
In contrast, our  warping layer is designed to be both differentiable and computationally efficient, which is necessary for online synthesis during the training.
\input{figures/figure4}

First, the live reconstruction of a spoof face $\mathbf{I}^i$ can be expressed as:
\begin{equation}
    G^i = G(\mathbf{I}^i)[\mathbf{p}_0],
    \label{eq:g0}
\end{equation}
where $\mathbf{p}^0=\{(0,0),(0,1),...,(255,255)\} \in \mathbb{R}^{256\times256\times2}$ enumerates pixel locations in $\mathbf{I}^i$.
To align the spoof traces while synthesizing a new spoof face, a dense offset $\Delta\mathbf{p}^{i\rightarrow j}\in \mathbb{R}^{256\times256\times2}$ is required to indicate the deformation between face $\mathbf{I}^i$ and face $\mathbf{I}^j$. 
A discrete deformation can be acquired from the distances of the corresponding facial landmarks between two faces. During the data preparation, we use \cite{dense-face-alignment} to fit a $3$DMM model and extract the $2$D locations of $Q$ facial vertices for each face:
\begin{equation}
    \mathbf{s}=\{(x_0,y_0),(x_1,y_1),...,(x_N,y_N)\} \in \mathbb{R}^{Q\times2}.
    \label{eq:s}
\end{equation}
A sparse offset on the corresponding vertices can then be computed two faces as $\Delta\mathbf{s}^{i\rightarrow j} = \mathbf{s}^j - \mathbf{s}^i$.
To convert the sparse offset $\Delta\mathbf{s}^{i\rightarrow j}$ to the dense offset $\Delta\mathbf{p}^{i\rightarrow j}$, we apply a triangulation interpolation:
\begin{equation}
    \Delta\mathbf{p}^{i\rightarrow j}=\text{Tri}(\mathbf{p}^0, \mathbf{s}^i,\Delta\mathbf{s}^{i\rightarrow j}),
    \label{eq:tri}
\end{equation}
where $\text{Tri}(\cdot)$ is the interpolation, $\mathbf{s}^i$ denotes the vertex locations, 
$\Delta\mathbf{s}^{i\rightarrow j}$ are the vertex values, and we adopt Delaunay triangulation. The warping operation can be denoted as:
\begin{equation}
    G^{-i\rightarrow j} = G^-(\mathbf{I}^j|\mathbf{I}^i)[\mathbf{p}^0+\Delta\mathbf{p}^{i\rightarrow j}],
    \label{eq:gx2y}
\end{equation}
where the offset $\Delta\mathbf{p}^{i\rightarrow j}$ applies to all subject $i$ related elements $\{\mathbf{T}_A^i, \mathbf{I}^i, \mathbf{P}^i\}$. Since the offset $\Delta\mathbf{p}^{i\rightarrow j}$  is typically composed of fractional numbers,  we implement the bilinear interpolation to sample the fractional pixel locations.
We select $Q=140$ vertices to cover the face region so that they can represent non-rigid deformation, due to pose and expression.
As the pixel values in the warped face are a linear combination of pixel values of the triangulation vertices, this entire process is differentiable. This process is illustrated in Fig.~\ref{fig:4}.

%% file: figures/figure4.tex
\begin{figure}[t!]
    \centering
    \resizebox{1\linewidth}{!}{\includegraphics{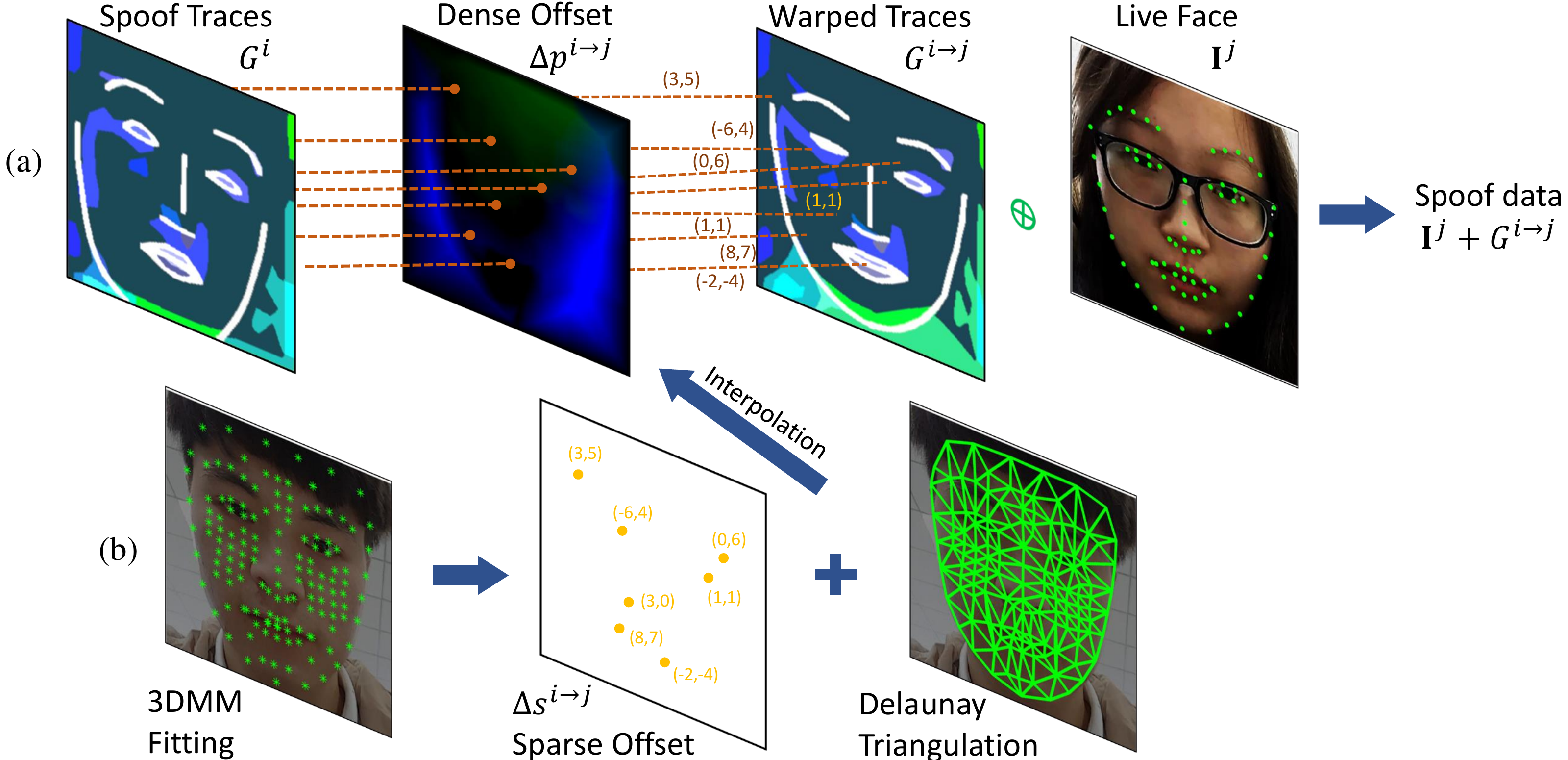}}
    \caption{\small The online $3$D warping layer. (a) Given the corresponding dense offset, we warp the spoof trace and add them to the target live face to create a new spoof. E.g. pixel $(x,y)$ with offset $(3,5)$ is warped to pixel$(x+3,y+5)$ in the new image. (b) To obtain a dense offsets from the spare offsets of the selected face shape vertices, Delaunay triangulation interpolation is adopted. }
    \label{fig:4}
\end{figure}

%% file: sec-3-3-dis.tex
\begin{algorithm*}[t]
\SetAlgoLined
\begin{flushleft}
    \textbf{Input:} live faces $\mathbf{I}_{\textit{live}}$ and facial landmarks $\mathbf{s}_{\textit{live}}$, spoof faces $\mathbf{I}_{\textit{spoof}}$ and facial landmarks $\mathbf{s}_{\textit{spoof}}$, ground truth depth map $\mathbf{M}_0$, preliminary mask $\mathbf{P}_0$;\\
    \textbf{Output:} reconstructed live $\hat{\mathbf{I}}_{\textit{live}}$, synthesized spoof $\hat{\mathbf{I}}_{\textit{spoof}}$, spoof traces $\{\mathbf{T}_A^{\textit{l}}$, $\mathbf{P}^{\textit{l}}$, $\mathbf{I}_P^{\textit{l}}, \mathbf{T}_A^{\textit{s}}$, $\mathbf{P}^{\textit{s}}$, $\mathbf{I}_P^{\textit{s}}\}$, depth maps $\{\mathbf{M}^{\textit{l}}$, $\mathbf{M}^{\textit{s}}\}$ ; 
    
\end{flushleft}
 \While{iteration $<$ max\_iteration}{
  \tcp{training step $1$}
  $\textbf{1}$: compute $\mathbf{T}_A^{\textit{l}}$, $\mathbf{P}^{\textit{l}}$, $\mathbf{I}_P^{\textit{l}}$ $\leftarrow$ $G(\mathbf{I}_{\textit{live}})$ and compute $\mathbf{T}_A^{\textit{s}}$, $\mathbf{P}^{\textit{s}}$, $\mathbf{I}_P^{\textit{s}}$ $\leftarrow$ $G(\mathbf{I}_{\textit{spoof}})$\;
  $\textbf{2}$: estimate the depth map $\mathbf{M}^{\textit{l}}$, $\mathbf{M}^{\textit{s}}$\;
  $\textbf{3}$: compute losses $L_{\textit{depth}}$, $L_P$, $L_R$\;
  
  \tcp{training step $2$}
  $\textbf{4}$: compute $\hat{\mathbf{I}}_{\textit{low}}$, $\hat{\mathbf{I}}_{\textit{mid}}$, $\hat{\mathbf{I}}_{\textit{hi}}$ from $\mathbf{T}_A^s$, $\mathbf{P}^s$, $\mathbf{I}_P^s$ and $\mathbf{I}_{\textit{spoof}}$ (Eqn.~\ref{eq:recon})\;
  $\textbf{5}$: compute warping offset $\Delta\mathbf{p}^{s\rightarrow l}$ from $\mathbf{s}_{\textit{live}}$, $\mathbf{s}_{\textit{spoof}}$ (Eqn.~\ref{eq:tri})\;
  $\textbf{6}$: compute $\hat{\mathbf{I}}_{\textit{spoof}}$ from warped $\mathbf{T}_A^{s\rightarrow l}$, $\mathbf{P}^{s\rightarrow l}$ and $\mathbf{I}_{\textit{live}}$ (Eqn.~\ref{eq:gx2y})\;
  $\textbf{7}$: send $\mathbf{I}_{\textit{live}}$, $\mathbf{I}_{\textit{spoof}}$, $\hat{\mathbf{I}}_{\textit{low}}$, $\hat{\mathbf{I}}_{\textit{mid}}$, $\hat{\mathbf{I}}_{\textit{hi}}$, $\hat{\mathbf{I}}_{\textit{spoof}}$ to discriminators\;
  $\textbf{8}$: compute the adversarial loss for generator $L_{\textit{G}}$ and for discriminators $L_{\textit{D}}$\;
  
  \tcp{training step $3$}
  $\textbf{9}$: create harder samples $\mathbf{I}_{\textit{hard}}$ from $\mathbf{T}_A^{s\rightarrow l}$, $\mathbf{P}^{s\rightarrow l}$ and $\mathbf{I}_{\textit{live}}$ with random perturbation on traces\;
  $\textbf{10}$: compute $\mathbf{T}_A^h$, $\mathbf{P}^h$, $\mathbf{I}_P^h$ $\leftarrow$ $G(\mathbf{I}_{\textit{hard}})$\;
  $\textbf{11}$: compute depth map $\mathbf{M}^h$ for $\mathbf{I}_{\textit{hard}}$\;
  $\textbf{12}$: compute losses $L_S$, $L_H$\;
  \tcp{back propagation}
  $\textbf{13}$: back-propagate the losses from step $3,8,12$ to corresponding parts and update the network;
 }
 \caption{PhySTD Training Iteration.}
 \label{alg:1}
\end{algorithm*}

\SubSection{Multi-scale Discriminators}
Motivated by~\cite{wang2018pix2pixHD}, we adopt multiple discriminators at different resolutions (\textit{e.g.}, $32$, $96$, and $256$) in our GAN architecture.
We follow the design of PatchGAN~\cite{isola2017image}, which essentially is a fully convolutional network. 
Fully convolutional networks are shown to be effective to not only synthesize high-quality images~\cite{isola2017image,wang2018pix2pixHD}, but also tackle face anti-spoofing problems~\cite{liu-auxiliary-fas}.
For each discriminator, we adopt the same structure but do not share the weights.  

As shown in Fig.~\ref{fig:3}, we use in total $4$ discriminators in our work: $D_1$, working in the lowest resolution of $32$, focuses on low frequency elements since the higher-frequency traces are erased by downsampling. 
$D_2$, working at the resolution of $96$, focuses on the middle level content pattern.
$D_3$ and $D_4$, working on the highest resolution of $256$, focus on the fine texture details.
Our preliminary version resizes real and synthetic samples $\{\mathbf{I}, \hat{\mathbf{I}}\}$ to different resolutions and assign to each discriminator.
To remove semantic ambiguity and provide correspondence to the trace components, we instead assign the hierarchical reconstruction from Eqn.~\ref{eq:recon_deco} to the discriminators: we send low frequency pairs $\{\mathbf{I}_{\textit{live}}, \hat{\mathbf{I}}_{\textit{low}}\}$ to $D_1$, middle frequency pairs $\{\mathbf{I}_{\textit{live}}, \hat{\mathbf{I}}_{\textit{mid}}\}$ to $D_2$, high frequency pairs $\{\mathbf{I}_{\textit{live}}, \hat{\mathbf{I}}_{\textit{hi}}\}$ to $D_3$, and real/synthetic spoof $\{\mathbf{I}_{\textit{spoof}}, \hat{\mathbf{I}}_{\textit{spoof}}\}$ to $D_4$.
Each discriminator outputs a $1$-channel map in the range of $[0,1]$, where $0$ denotes fake and $1$ denotes real. 

%% file: sec-3-4-loss.tex
\SubSection{Loss Functions and Training Steps}
\label{sec:3-loss}
We utilize multiple loss functions to supervise the learning of depth maps and spoof traces. Each training iteration consists of three training steps. We first introduce the loss function, followed by how they are used in the training steps.

\Paragraph{Depth map loss:} 
We follow the auxiliary FAS~\cite{liu-auxiliary-fas} to estimate an auxiliary depth map $\mathbf{M}$, where the depth ground truth $\mathbf{M}_0$ for a live face contains face-like shape and the depth for spoof should be zero.
We apply the $\mathcal{L}$-$1$ norm on this loss as:
\begin{equation}
    L_{\textit{depth}} = \frac{1}{K^2}\mathbb{E}_{i\sim \mathcal{L}\cup\mathcal{S}}
    {\lVert}\mathbf{M}^i-\mathbf{M}^i_0{\rVert}_F, 
    \label{eq:dloss}
\end{equation}
where $K\!=\!32$ is the size of $\mathbf{M}$. We apply the dense face alignment~\cite{dense-face-alignment} to estimate the $3$D shape and render the depth ground truth $\mathbf{M}_0$.
\input{figures/figure4new}

\Paragraph{Adversarial loss for $G$:} 
We employ the LSGANs~\cite{mao2017least} on reconstructed live faces and synthesized spoof faces. It encourages the reconstructed live to look similar to real live from domain $\mathcal{L}$, and the synthesized spoof faces to look similar to faces from domain $\mathcal{S}$: 
\begin{equation}
    \begin{aligned}
    L_{G} = \mathbb{E}_{i\sim \mathcal{L}, j\sim \mathcal{S}}\Big[
         {\lVert}D_1(\hat{\mathbf{I}}_{\textit{low}}^j)\!- \!\mathbf{1}{\rVert}_F^2 + {\lVert}D_2(\hat{\mathbf{I}}_{\textit{mid}}^j)\!- \!\mathbf{1}{\rVert}_F^2 + \\
         {\lVert}D_3(\hat{\mathbf{I}}_{\textit{hi}}^j)\!- \!\mathbf{1}{\rVert}_F^2 + 
         {\lVert}D_4(\hat{\mathbf{I}}_{\textit{spoof}}^{j\rightarrow i})\! -\mathbf{1}{\rVert}_F^2\Big].
    \end{aligned}
    \label{eq:gloss}
\end{equation}

\Paragraph{Adversarial loss for $D$:} 
The adversarial loss for discriminators encourages $D(\cdot)$ to distinguish between real live \textit{vs.}~reconstructed live, and real spoof \textit{vs.}~synthesized spoof:
\begin{equation}
\begin{aligned}
 L_{D} = \mathbb{E}_{i\sim \mathcal{L}, j\sim \mathcal{S}}\Big[
        {\lVert}D_1(\mathbf{I}^i)\!- \!\mathbf{1}{\rVert}_F^2 + 
        {\lVert}D_2(\mathbf{I}^i)\!- \!\mathbf{1}{\rVert}_F^2 + \\
        {\lVert}D_3(\mathbf{I}^i)\!- \!\mathbf{1}{\rVert}_F^2 +
        {\lVert}D_4(\mathbf{I}^j)\! -\mathbf{1}{\rVert}_F^2 + 
        {\lVert}D_1(\hat{\mathbf{I}}_{\textit{low}}^j){\rVert}_F^2 + \\
        {\lVert}D_2(\hat{\mathbf{I}}_{\textit{mid}}^j){\rVert}_F^2 +
        {\lVert}D_3(\hat{\mathbf{I}}_{\textit{hi}}^j){\rVert}_F^2 + {\lVert}D_4(\hat{\mathbf{I}}_{\textit{spoof}}^{j\rightarrow i}){\rVert}_F^2\Big].
\end{aligned}
\label{eq:dloss}
\end{equation}

\Paragraph{Inpainting mask loss:} 
The ground truth inpainting region for all spoof attacks is barely possible to obtain, hence  a fully supervised training~\cite{missing-modalities-imputation-via-cascaded-residual-autoencoder} for inpainting mask is out of the question.
However, we may still leverage the prior knowledge of spoof attacks to facilitate the estimation of inpainting masks.
The inpainting mask loss consists of a positive term and a negative term. 
First, the positive term encourages certain region to be inpainted.
As the goal of inpainting process is to allow certain region to change without intensity constraint, the region with larger magnitude of additive traces would have a higher probability to be inpainted.
Hence, the positive term adopts a $\mathcal{L}$-$2$ norm between the inpainting region $\mathbf{P}$ and the region where the additive trace is larger than a threshold $\beta$.

Second, the negative term discourages certain region to be inpainted. While the ground truth inpainting mask is unknown, it's straightforward to mark a large portion of region that should not be inpainted. For instance, the inpainting region for funny eye glasses should not appear in the lower part of a face. 
Hence, we provide a preliminary mask $\mathbf{P}_0$ to indicate the not-to-be-inpainted region, and adopt a normalized $\mathcal{L}$-$2$ norm on the masked inpainting region $\mathbf{P}\cdot\mathbf{P}_0$ as the negative term.
The preliminary mask $\mathbf{P}_0$ is illustrated in Fig.~\ref{fig:aug}.
Overall, the inpainting mask loss is formed as:

\begin{equation}
    L_{P} = \mathbb{E}_{\mathbf{i}\sim \mathcal{S}}\Big[
    {\lVert}\mathbf{P}^i-(\mathbf{T}_A^i>\beta){\rVert}_F^2 +
    \frac{{\lVert}\mathbf{P}^i\cdot\mathbf{P}_0^i{\rVert}_F^2 }{{\lVert}\mathbf{P}_0^i{\rVert}_F^2}
     \Big].
    \label{eq:ploss}
\end{equation}

\Paragraph{Trace regularization:} 
Based on Eqn.~\ref{eq:op} with $\lambda=1$, we regularize the intensity of additive traces $\{\mathbf{B},\mathbf{C},\mathbf{T}\}$ and inpainting region $\mathbf{P}$. The regularizer loss is denoted as: 
\begin{equation}
    L_{R} = \mathbb{E}_{\mathbf{i}\sim \mathcal{L}\cup\mathcal{S}}\Big[
    {\lVert}\mathbf{B}{\rVert}_F^2 + 
    {\lVert}\mathbf{C}{\rVert}_F^2 + 
    {\lVert}\mathbf{T}{\rVert}_F^2 + 
    {\lVert}\mathbf{P}{\rVert}_F^2\Big].
    \label{eq:rloss}
\end{equation}

\Paragraph{Synthesized spoof loss:} 
Synthesized spoof data come with ground truth spoof traces. As a result, we are able to define a supervised pixel loss for the generator to disentangle the exact spoof traces that were added:
\begin{equation} 
    L_{S} = \mathbb{E}_{\mathbf{i}\sim \mathcal{L}, \mathbf{j}\sim \mathcal{S}}
    \Big[{\lVert}G(\lceil G^{-j\rightarrow i}  \rceil) - 
    \lceil G^{j\rightarrow i}\rceil {\rVert}_F^1\Big],
    \label{eq:sloss}
\end{equation}
where $G^{j\rightarrow i}$ is the overall effect of $\{\mathbf{P}^j,\mathbf{I}_P^j,\mathbf{B}^j,\mathbf{C}^j,\mathbf{T}^j\}$ after warping to subject $i$, and $\lceil \cdot \rceil$ is the \verb|stop_gradient| operation.
Without stopping the gradient, $G^{j\rightarrow i}$ may collapse to $0$.

\Paragraph{Depth map loss for ``harder'' samples:} 
We send the ``harder'' synthesized spoof data to depth estimation network to improve the data diversity, and hope to increase the FAS model's generalization:
\begin{equation}
    L_{\textit{H}} = \frac{1}{K^2}\mathbb{E}_{i\sim \hat{\mathcal{S}}}
    \Big[{\lVert}\mathbf{M}^i-\mathbf{M}_0^i{\rVert}_F\Big],
    \label{eq:hloss}
\end{equation}
where $\hat{\mathcal{S}}$ denotes the domain of synthesized spoof faces. 

\Paragraph{Training steps and total loss:} 
Each training iteration has $3$ training steps.
In the training step $1$, live faces $\mathbf{I}_{\textit{live}}$ and spoof faces $\mathbf{I}_{\textit{spoof}}$ are fed into generator $G(\cdot)$ to disentangle the spoof traces. The spoof traces are used to reconstruct the live counterpart $\hat{\mathbf{I}}_{\textit{live}}$ and synthesize new spoof $\hat{\mathbf{I}}_{\textit{spoof}}$. The generator is updated with respect to the depth map loss $L_{\textit{depth}}$, adversarial loss $L_G$,  inpainting mask loss $L_P$, and regularizer loss $L_R$:
\begin{equation}
    L =  \alpha_1L_{\textit{depth}} + \alpha_2 L_{G} + \alpha_3 L_P + \alpha_4 L_R.
    \label{eq:gloss_overall}
\end{equation}

In the training step $2$, the discriminators are supervised with the adversarial loss $L_D$ to compete with the generator.
In the training step $3$, $\mathbf{I}_{\textit{live}}$ and $\hat{\mathbf{I}}_{\textit{hard}}$ are fed into the generator with the ground truth label and trace to minimize the synthesized spoof loss $L_S$ and depth map loss $L_H$: 
\begin{equation}
    L = \alpha_5 L_{S} + \alpha_6L_H,
    \label{eq:aloss_overall}
\end{equation}
where $\alpha_1$-$\alpha_6$ are the weights to balance the multitask training. 
To note that, we send the original live faces $\mathbf{I}_{\textit{live}}$ with $\hat{\mathbf{I}}_{\textit{hard}}$ for a balanced mini-batch, which is important when computing the moving average in the batch normalization layer.
We execute all $3$ steps in each minibatch iteration, but reduce the learning rate for discriminator step by half.
The whole training process is depicted in Alg.~\ref{alg:1}.

%% file: figures/figure4new.tex
\begin{figure}[t!]
    \centering
    \resizebox{1\linewidth}{!}{\includegraphics{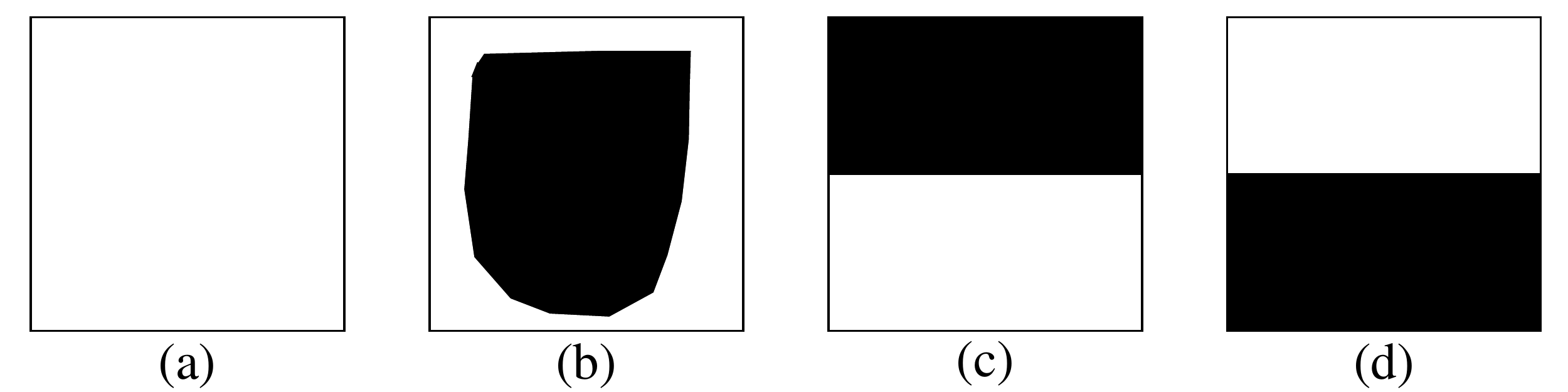}}
    \caption{\small Preliminary mask $\mathbf{P}_0$ for the negative term in inpainting mask loss. White pixels denote $1$ and black pixels denote $0$. White  indicates the area should not be inpainted.  $\mathbf{P}_0$ for: (a) print, replay; (b) 3D mask and makeup; (c) partial attacks that cover the eye portion; (d) partial attacks that cover the mouth portion. }
    \label{fig:aug}
\end{figure}

%% file: sec-4-0-main.tex
\Section{Experiments}
\label{sec:4}
In this section, we first introduce the experimental setup, and then present the results in the known, unknown, and open-set spoof scenarios, with comparisons to respective baselines.
Next, we quantitatively evaluate the spoof traces by performing a spoof medium classification, and conduct an ablation study on each design in the proposed method.
Finally, we provide visualization results on the spoof trace disentanglement, new spoof synthesis and t-SNE visualization.
\input{figures/table-known}

\input{sec-4-1-setup}

\input{sec-4-2-known}

\input{sec-4-3-unknown}

\input{sec-4-4-trace-cls}
\input{sec-4-5-ablation}
\input{sec-4-6-vis}

%% file: figures/table-known.tex
\begin{table}[t!]
\small\centering
\caption{\small The evaluation on four protocols in OULU-NPU. \textbf{Bold} indicates the best score in each protocol.}
	
\resizebox{0.95\textwidth}{!}{
\begin{tabular}{cllll}
    \toprule Protocol & Method & APCER (\%) & BPCER (\%) & ACER (\%) \\ \hline
    \multirow{8}{*}{1} 
           & STASN\cite{yang2019face} & $1.2$ & $2.5$ & $1.9$ \\
           & Auxiliary~\cite{liu-auxiliary-fas} & $1.6$ & $1.6$ & $1.6$ \\
           & DeSpoof~\cite{jourabloo-face-despoofing} & $1.2$ & $1.7$ & $1.5$ \\
           & DRL~\cite{drl2020} & $1.7$ & $0.8$ & $1.3$ \\
           & STDN~\cite{liu2020on} & $0.8$ & $1.3$ & $1.1$ \\
           & CDCN~\cite{cdcn} & $0.4$ & $1.7$ & $1.0$ \\
           & HMP~\cite{hmp2020} & $0.0$ & $1.6$ & $0.8$ \\
           & CDCN++~\cite{cdcn} & $0.4$ & $\textbf{0.0}$ & $\textbf{0.2}$ \\
           & Ours & $\textbf{0.0}$ & $0.8$ & $0.4$ \\ \hline
    \multirow{8}{*}{2} 
           & DeSpoof~\cite{jourabloo-face-despoofing} & $4.2$ & $4.4$ & $4.3$ \\
           & Auxiliary~\cite{liu-auxiliary-fas} & $2.7$ & $2.7$ & $2.7$ \\
           & DRL~\cite{drl2020} & $1.1$ & $3.6$ & $2.4$ \\
           & STASN\cite{yang2019face} & $4.2$ & $\textbf{0.3}$ & $2.2$ \\
           & STDN~\cite{liu2020on} & $2.3$ & $1.6$ & $1.9$ \\
           & HMP~\cite{hmp2020} & $2.6$ & $0.8$ & $1.7$ \\
           & CDCN~\cite{cdcn} & $0.4$ & $1.7$ & $1.5$ \\
           & CDCN++~\cite{cdcn} & $1.8$ & $0.8$ & $1.3$ \\
           & Ours & $\textbf{1.2}$ & $1.3$ & $\textbf{1.3}$ \\ \hline
    \multirow{8}{*}{3} 
           & DeSpoof~\cite{jourabloo-face-despoofing} & $4.0\pm1.8$ & $3.8\pm1.2$ & $3.6\pm1.6$ \\
           & Auxiliary~\cite{liu-auxiliary-fas} & $2.7\pm1.3$ & $3.1\pm1.7$ & $2.9\pm1.5$ \\
           & STDN~\cite{liu2020on} & $1.6\pm1.6$ & $4.0\pm5.4$ & $2.8\pm 3.3$ \\
           & STASN\cite{yang2019face} & $4.7\pm3.9$ & $\textbf{0.9}\pm\textbf{1.2}$ & $2.8\pm1.6$ \\
           & HMP~\cite{hmp2020} & $2.8\pm2.4$ & $2.3\pm2.8$ & $2.5\pm 1.1$ \\
           & CDCN~\cite{cdcn} & $2.4\pm1.3$ & $2.2\pm2.0$ & $2.3\pm 1.4$ \\
           & DRL~\cite{drl2020} & $2.8\pm2.2$ & $1.7\pm2.6$ & $2.2\pm 2.2$ \\
           & CDCN++~\cite{cdcn} & $1.7\pm1.5$ & $2.0\pm1.2$ & $\textbf{1.8}\pm \textbf{0.7}$ \\
           & Ours & $\textbf{1.7}\pm \textbf{1.4}$ & $2.2\pm 3.5$ & $1.9\pm 2.3$ \\ \hline
    \multirow{8}{*}{4} 
           & Auxiliary~\cite{liu-auxiliary-fas} & $9.3\pm5.6$ & $10.4\pm6.0$ & $9.5\pm6.0$ \\
           & STASN\cite{yang2019face} & $6.7\pm10.6$ & $8.3\pm8.4$ & $7.5\pm4.7$ \\
           & CDCN~\cite{cdcn} & $4.6\pm4.6$ & $9.2\pm8.0$ & $6.9\pm2.9$ \\
           & DeSpoof~\cite{jourabloo-face-despoofing} & $5.1\pm6.3$ & $6.1\pm5.1$ & $5.6\pm5.7$ \\
           & HMP~\cite{hmp2020} & $2.9\pm4.0$ & $7.5\pm6.9$ & $5.2\pm 3.7$ \\
           & CDCN++~\cite{cdcn} & $4.2\pm3.4$ & $5.8\pm4.9$ & $5.0\pm 2.9$ \\
           & DRL~\cite{drl2020} & $5.4\pm2.9$ & $\textbf{3.3}\pm\textbf{6.0}$ & $4.8\pm 6.4$ \\
           & STDN~\cite{liu2020on} & $2.3\pm3.6$ & $5.2\pm5.4$ & $3.8\pm 4.2$ \\
           & Ours & $\textbf{2.3}\pm \textbf{3.6}$ & $4.2\pm 5.4$ & $\textbf{3.6}\pm \textbf{4.2}$ \\ \bottomrule  
           \tiny 
\end{tabular}}
\label{tab:known1}
\end{table}

%% file: sec-4-1-setup.tex
\SubSection{Experimental Setup}
\Paragraph{Databases}
We conduct experiments on three major databases: Oulu-NPU~\cite{OULU_NPU_2017}, SiW~\cite{liu-auxiliary-fas}, and SiW-M~\cite{deep-tree}.
Oulu-NPU and SiW include print/replay attacks, while SiW-M includes $13$ spoof types.
We follow all the existing testing protocols and compare with SOTA methods.
Similar to most prior works, we only use the face region for training and testing.

\Paragraph{Evaluation metrics}
Two common metrics are used in this work for comparison: EER and APCER/BPCER/ACER.
EER describes the theoretical performance and predetermines the threshold for making decisions.
APCER/BPCER/ACER\cite{acer1} describe the practical performance given a predetermined threshold.
For both evaluation metrics, lower value means better performance.
The threshold for APCER/BPCER/ACER is computed from either training set or validation set.
In addition, we also report the True Detection Rate (TDR) at a given False Detection Rate (FDR). This metric describes the spoof detection rate at a strict tolerance to live errors, which is widely used to evaluate real-world systems~\cite{iarpa-odin}. In this work, we report TDR at FDR$=0.5\%$.
For TDR, the higher the better.

\Paragraph{Parameter setting}
PhySTD is implemented in Tensorflow with an initial learning rate of $5e$-$5$.
We train in total $150,000$ iterations with a batch size of $8$, and decrease the learning rate by a ratio of $10$ every $45,000$ iterations. 
We initialize the weights with $[0,0.02]$ normal distribution. 
$\{\alpha_1,\alpha_2,\alpha_3,\alpha_4,\alpha_5,\alpha_6\}$ are set to be $\{100,5,1,1e$-$4,10,1\}$, and $\beta=0.1$.
$\alpha_0$ is empirically determined from the training or validation set.
We use the open-source face alignment~\cite{bulat2017far} and $3$DMM fitting~\cite{dense-face-alignment} to crop the face and provide $140$ landmarks.
\input{figures/table-known-2}

%% file: figures/table-known-2.tex
\begin{table}[t!]
\small
\centering
\caption{\small The evaluation on three protocols in SiW Dataset. We compare with the top $7$ performances.}
	
\resizebox{0.95\textwidth}{!}{
\begin{tabular}{cllll}
    \toprule Protocol & Method & APCER (\%) & BPCER (\%) & ACER (\%) \\ \hline
    \multirow{8}{*}{1} 
        & Auxiliary\cite{liu-auxiliary-fas} & $3.6$ & $3.6$ & $3.6$ \\
        & STASN\cite{yang2019face} & $-$ & $-$ & $1.0$ \\
        & Meta-FAS-DR\cite{zhao2019meta} & $0.5$ & $0.5$ & $0.5$ \\
        & HMP~\cite{hmp2020} & $0.6$ & $0.2$ & $0.5$ \\
        & DRL~\cite{drl2020} & $0.1$ & $0.5$ & $0.3$ \\
        & CDCN~\cite{cdcn} & $0.1$ & $0.2$ & $0.1$ \\
        & CDCN++~\cite{cdcn} & $0.1$ & $0.2$ & $0.1$ \\
        & Ours & $\textbf{0.0}$ & $\textbf{0.0}$ & $\textbf{0.0}$ \\ \hline
    \multirow{8}{*}{2} 
        & Auxiliary\cite{liu-auxiliary-fas} & $0.6\pm0.7$ & $0.6\pm0.7$ & $0.6\pm0.7$ \\
        & Meta-FAS-DR\cite{zhao2019meta} & $0.3\pm0.3$ & $0.3\pm0.3$ & $0.3\pm0.3$ \\
        & STASN\cite{yang2019face} & $-$ & $-$ & $0.3\pm0.1$ \\
        & HMP~\cite{hmp2020} & $0.1\pm 0.2$ & $0.2\pm 0.0$ & $0.1\pm 0.1$ \\
        & DRL~\cite{drl2020} & $0.1\pm 0.2$ & $0.1\pm 0.1$ & $0.1\pm 0.0$ \\
        & CDCN~\cite{cdcn} & $0.0\pm0.0$ & $0.1\pm0.1$ & $0.1\pm 0.0$ \\
        & CDCN++~\cite{cdcn} & $0.0\pm0.0$ & $0.1\pm0.1$ & $0.0\pm 0.1$ \\
        & Ours & $\textbf{0.0}\pm\textbf{0.0}$ & $\textbf{0.0}\pm\textbf{0.0}$ & $\textbf{0.0}\pm\textbf{0.0}$ \\ \hline
    \multirow{8}{*}{3} 
        & STASN\cite{yang2019face} & $-$ & $-$ & $12.1\pm1.5$ \\
        & Auxiliary\cite{liu-auxiliary-fas} & $8.3\pm3.8$ & $8.3\pm3.8$ & $8.3\pm3.8$ \\
        & Meta-FAS-DR\cite{zhao2019meta} & $8.0\pm5.0$ & $7.4\pm5.7$ & $7.7\pm5.3$ \\
        & DRL~\cite{drl2020} & $9.4\pm 6.1$ & $1.8\pm 2.6$ & $5.6\pm 4.4$ \\
        & HMP~\cite{hmp2020} & $2.6\pm 0.9$ & $2.3\pm 0.5$ & $2.5\pm 0.7$ \\
        & CDCN~\cite{cdcn} & $2.4\pm1.3$ & $2.2\pm2.0$ & $2.3\pm 1.4$ \\
        & CDCN++~\cite{cdcn} & $\textbf{1.7}\pm\textbf{1.5}$ & $\textbf{2.0}\pm\textbf{1.2}$ & $\textbf{1.8}\pm \textbf{0.7}$ \\
        & Ours & $13.1\pm 9.4$ & $1.6\pm 0.6$ & $7.4\pm 4.3$ \\ \bottomrule
        \tiny 
\end{tabular}}
\label{tab:known2}
\end{table}

%% file: sec-4-2-known.tex
\SubSection{Anti-Spoofing for Known Spoof Types}

\Paragraph{Oulu-NPU}
Oulu-NPU\cite{OULU_NPU_2017} is a commonly used face anti-spoofing benchmark due to its high-quality data and challenging testing protocols.
Tab.~\ref{tab:known1} shows our anti-spoofing performance on Oulu-NPU, compared with SOTA algorithms.
Our method achieves the best overall performance on this database.
Compared with our preliminary version~\cite{liu2020on}, we demonstrate improvements in all $4$ protocols, with significant improvement on protocol $1$ and protocol $3$, {\it i.e.}, reducing the ACER by $63.6\%$ and $32.1\%$ respectively.
Compared with the SOTA, our approach achieves similar best performances on the first three protocols and outperforms the SOTA on the fourth protocol, which is the most challenging one. To note that, in protocol $3$ and protocol $4$, the performances of testing camera $6$ are much lower than those of cameras $1$-$5$: the ACER for camera $6$ are $6.4\%$ and $10.2\%$, while the average ACER for the other cameras are $1.0\%$ and $2.0\%$ respectively. 
Compared with other cameras, we notice that camera $6$ has stronger sensor noises and our model recognizes them as unknown spoof traces, which leads to an increased false negative rate (\textit{i.e.}, BPCER).
How to separate sensor noises from spoof traces can be an important future research topic. 
\input{figures/table-siwm-p1}
\input{figures/table-siwm-p2}

\Paragraph{SiW}
SiW\cite{liu-auxiliary-fas} is another recent high-quality database.
It includes fewer capture cameras but more spoof mediums and environment variations, such as pose, illumination, and expression.
The comparison on three protocols is shown in Tab.~\ref{tab:known2}.
We outperform the previous works on the first two protocols and rank in the middle on protocol $3$. 
Protocol $3$ aims to test the performance of unknown spoof detection, where the model is trained on one spoof attack (print or replay) and tested on the other.
As we can see from Fig.\ref{fig:10}-\ref{fig:6}, the traces of print and replay are significantly different, where the replay traces are more on the high-frequency part (\textit{i.e.}, trace component $\textbf{T}$) and the print traces are more on the low-frequency part (\textit{i.e.}, trace component $\textbf{S}$). 
These pattern divergence leads to the adaption gap of our method while training on one attack and testing on the other.

\Paragraph{SiW-M}
SiW-M\cite{deep-tree} contains a large diversity of spoof types, including print, replay, $3$D mask, makeup, and partial attacks.
This allows us to have a comprehensive evaluation of the proposed approach with different spoof attacks.
To use SiW-M for known spoof detection, we randomly split the data of all types into train/test set with a ratio of $60\%$ vs.~$40\%$, and the results are shown in Tab.~\ref{tab:siwm_p1}.
Compared to the preliminary version~\cite{liu2020on}, our method outperforms on most spoof types as well as the overall EER performance by $47.9\%$ relatively, 
which demonstrates the superiority of our anti-spoofing on known spoof attacks.

For experiments on SiW-M (protocol I, II, and III), we additionally report the TPR at FNR equal to $0.5\%$.
While EER and ACER provide the theoretical evaluation, the users in real-world applications care more about the true spoof detection rate under a given live detection error rate, and hence TPR can better reflect how well the model can detect one or a few spoof attacks in practices. 
As shown in Tab.~\ref{tab:siwm_p1}, we improve the overall TDR of our preliminary version~\cite{liu2020on} by $29.5\%$.

%% file: figures/table-siwm-p1.tex
\begin{table*}[t]
\caption{\small The evaluation and ablation study on SiW-M Protocol I: known spoof detection.}
\centering
	\resizebox{1\textwidth}{!}{
	\begin{tabular}{rllllllllllllllllll}
	\multicolumn{17}{c}{} \\ \toprule
	\multirow{2}{*}{Metrics(\%)} & \multirow{2}{*}{Method} & \multirow{2}{*}{Replay}& \multirow{2}{*}{Print} & \multicolumn{5}{c}{3D Mask}  && \multicolumn{3}{c}{Makeup}  &&  \multicolumn{3}{c}{Partial Attacks} & \multirow{2}{*}{Overall}\\ 
	\cline{5-9} \cline{11-13} \cline{15-17}
	& & & & Half & Silic. & Trans. & Paper & Mann. && Ob. & Im. & Cos. && Funny. & Papergls. & Paper &\\ \hline
    \multirow{6}{*}{ACER} & Auxiliary\cite{liu-auxiliary-fas} 
        &$5.1$& $5.0$ & $5.0$ & $10.2$ & $5.0$ & $9.8$ & $6.3$ && $19.6$ & $5.0$ & $26.5$ && $5.5$ & $5.2$ & $5.0$ & $6.3$ \\
    & SDTN\cite{liu2020on} 
        &$3.2$& $3.1$ & $3.0$ & $9.0$ & $3.0$ & $3.4$ & $4.7$ && $\textbf{3.0}$ & $3.0$ & $24.5$ && $4.1$ & $3.7$ & $3.0$ & $4.1$ \\ 
    & Step$1$ 
        &$6.1$& $5.4$ & $5.4$ & $5.4$ & $5.4$ & $5.4$ & $5.4$ && $22.7$ & $5.4$ & $26.8$ && $5.4$ & $5.5$ & $5.4$ & $10.9$ \\
    & Step$1$+Step$2$ w/ single trace
        &$8.7$& $7.8$ & $7.8$ & $7.8$ & $7.8$ & $7.8$ & $7.8$ && $25.0$ & $7.9$ & $28.8$ && $7.8$ & $7.8$ & $7.8$ & $13.8$ \\
    & Step$1$+Step$2$
        &$4.1$& $3.9$ & $3.9$ & $3.9$ & $4.0$ & $3.9$ & $4.0$ && $13.5$ & $4.0$ & $25.1$ && $3.9$ & $3.9$ & $3.9$ & $4.6$ \\
    & Step$1$+Step$2$+Step$3$ (Ours) 
        &$\textbf{3.2}$& $\textbf{1.4}$ & $\textbf{1.0}$ & $\textbf{2.3}$ & $\textbf{1.3}$ & $\textbf{2.9}$ & $\textbf{2.5}$ && $12.4$ & $\textbf{1.2}$ & $\textbf{18.5}$ && $\textbf{1.7}$ & $\textbf{0.4}$ & $\textbf{1.6}$ & $\textbf{2.8}$ \\\hline
    \multirow{6}{*}{EER} & Auxiliary\cite{liu-auxiliary-fas}
        &$4.7$& $0.0$ & $1.6$ & $10.5$ & $4.6$ & $10.0$ & $6.4$ && $12.7$ & $0.0$ & $19.6$ && $9.3$ & $7.5$ & $0.0$ & $6.7$ \\
    & SDTN\cite{liu2020on}
        &$2.1$& $2.2$ & $0.0$ & $7.2$ & $0.1$ & $3.9$ & $4.8$ && $\textbf{0.0}$ & $0.0$ & $19.6$ && $5.3$ & $5.4$ & $\textbf{0.0}$ & $4.8$ \\
    & Step$1$ 
        &$3.8$& $2.7$ & $1.5$ & $2.7$ & $1.9$ & $\textbf{1.8}$ & $2.4$ && $15.1$ & $0.7$ & $28.7$ && $4.1$ & $4.9$ & $1.0$ & $4.3$ \\
    & Step$1$+Step$2$ w/ single trace
        &$6.7$& $5.3$ & $0.8$ & $\textbf{1.5}$ & $1.4$ & $3.3$ & $3.2$ && $21.5$ & $1.0$ & $27.1$ && $6.5$ & $6.1$ & $1.5$ & $5.8$ \\
    & Step$1$+Step$2$ 
        &$2.4$& $3.1$ & $0.4$ & $2.6$ & $1.2$ & $3.0$ & $2.4$ && $9.5$ & $0.4$ & $23.5$ && $1.1$ & $0.5$ & $0.6$ & $2.8$ \\
    & Step$1$+Step$2$+Step$3$ (Ours)
        &$\textbf{2.5}$& $\textbf{1.0}$ & $\textbf{0.0}$ & $2.1$ & $\textbf{1.0}$ & $1.9$ & $\textbf{2.2}$ && $8.2$ & $\textbf{0.0}$ & $\textbf{18.5}$ && $\textbf{0.8}$ & $\textbf{0.0}$ & $0.4$ & $\textbf{2.5}$ \\ \hline
     & SDTN\cite{liu2020on}
        &$\textbf{90.1}$& $76.1$ & $80.7$ & $71.5$ & $62.3$ & $74.4$ & $85.0$ && $\textbf{100}$ & $100$ & $33.8$ && $49.6$ & $30.6$ & $97.7$ & $70.4$ \\
    TPR@ & Step$1$
        &$43.8$& $43.3$ & $47.2$ & $44.5$ & $62.9$ & $54.8$ & $55.4$ && $16.7$ & $90.6$ & $31.5$ && $60.3$ & $56.7$ & $77.1$ & $59.3$ \\
    FNR=$.5\%$ & Step$1$+Step$2$ w/ single trace
        &$58.9$& $76.8$ & $97.6$ & $\textbf{94.2}$ & $94.9$ & $66.3$ & $78.3$ && $13.3$ & $94.1$ & $49.1$ && $62.4$ & $58.5$ & $92.1$ & $74.8$ \\
    & Step$1$+Step$2$
        &$84.7$& $74.7$ & $100$ & $70.1$ & $\textbf{96.6}$ & $77.5$ & $89.6$ && $36.9$ & $100$ & $40.1$ && $96.3$ & $99.4$ & $99.4$ & $89.7$ \\
    & Step$1$+Step$2$+Step$3$ (Ours)
        &$85.7$& $\textbf{85.4}$ & $\textbf{100}$ & $76.6$ & $96.3$ & $\textbf{80.2}$ & $\textbf{93.8}$ && $41.1$ & $\textbf{100}$ & $\textbf{55.8}$ && $\textbf{98.1}$ & $\textbf{100}$ & $\textbf{99.8}$ & $\textbf{91.2}$ \\\bottomrule 
	\end{tabular}
\label{tab:siwm_p1}} 
\end{table*}

%% file: figures/table-siwm-p2.tex
\begin{table*}[t!]
\small
\centering
\caption{\small The evaluation on SiW-M Protocol II: unknown spoof detection.}
\resizebox{\textwidth}{!}{
	\begin{tabular}{rlllllllllllllllll}
	\multicolumn{15}{c}{} \\ \toprule
	Metrics & \multirow{2}{*}{Method} &\multirow{2}{*}{Replay}& \multirow{2}{*}{Print} & \multicolumn{5}{c}{3D Mask}  && \multicolumn{3}{c}{Makeup}  &&  \multicolumn{3}{c}{Partial Attacks} & \multirow{2}{*}{Average}\\
	\cline{5-9} \cline{11-13} \cline{15-17}
	(\%) & & & & Half & Silic. & Trans. & Paper & Mann. && Ob. & Im. & Cos. && Fun. & Papergls. & Paper &\\ \hline
    \multirow{8}{*}{APCER} & Auxiliary\cite{liu-auxiliary-fas}
        &$23.7$& $7.3$ & $27.7$ & $18.2$ & $97.8$ & $8.3$ & $16.2$ && $100.0$ & $18.0$ & $16.3$ && $91.8$ & $72.2$ & $0.4$ & $38.3\pm37.4$  \\ 
    & LBP+SVM~\cite{OULU_NPU_2017}
        &$19.1$& $15.4$ & $40.8$ & $20.3$ & $70.3$ & $0.0$ & $4.6$ && $96.9$ & $35.3$ & $\mathbf{11.3}$ && $53.3$ & $58.5$ & $0.6$ & $32.8\pm29.8 $ \\ 
    & DTL\cite{deep-tree}
        &$\textbf{1.0}$& $0.0$ & $0.7$ & $24.5$ & $58.6$ & $0.5$ & $3.8$ && $73.2$ & $13.2$ & $12.4$ && $17.0$ & $17.0$ & $0.2$ & $17.1\pm23.3$  \\ 
    & CDCN \cite{cdcn}
        &$8.2$& $6.9$ & $8.3$ & $7.4$ & $20.5$ & $5.9$ & $5.0$ && $43.5$ & $1.6$ & $14.0$ && $24.5$ & $18.3$ & $1.2$ & $12.7\pm11.7$ \\ 
    & SDTN\cite{liu2020on}
        &$1.6$& $\mathbf{0.0}$ & $\mathbf{0.5}$ & $\mathbf{7.2}$ & $9.7$ & $0.5$ & $\mathbf{0.0}$ && $96.1$ & $0.0$ & $21.8$ && $\mathbf{14.4}$ & $\mathbf{6.5}$ & $0.0$ & $12.2\pm26.1$  \\
    & CDCN++\cite{cdcn}
        &$9.2$& $6.0$ & $4.2$ & $7.4$ & $18.2$ & $\mathbf{0.0}$ & $5.0$ && $39.1$ & $0.0$ & $14.0$ && $23.3$ & $14.3$ & $0.0$ & $10.8\pm\ 11.2$ \\ 
    & HMP\cite{hmp2020}
        &$12.4$& $5.2$ & $8.3$ & $9.7$ & $13.6$ & $\mathbf{0.0}$ & $2.5$ && $\mathbf{30.4}$ & $0.0$ & $12.0$ && $22.6$ & $15.9$ & $1.2$ & $\mathbf{10.3}\pm\mathbf{9.1}$ \\ 
    & Ours
        &$10.0$& $4.9$ & $5.3$ & $16.7$ & $\mathbf{3.5}$ & $2.0$ & $2.8$ && $92.8$ & $\mathbf{0.0}$ & $37.5$ && $33.7$ & $23.2$ & $0.2$ & $17.9\pm25.8$ \\ \hline
    \multirow{8}{*}{BPCER} & LBP+SVM~\cite{OULU_NPU_2017}
     &$22.1$& $21.5$ & $21.9$ & $21.4$ & $20.7$ & $23.1$ & $22.9$ && $21.7$ & $12.5$ & $22.2$ && $18.4$ & $20.0$ & $22.9$ & $21.0\pm2.9$  \\ 
    & DTL\cite{deep-tree}
        &$18.6$& $11.9$ & $29.3$ & $12.8$ & $13.4$ & $8.5$ & $23.0$ && $11.5$ & $9.6$ & $16.0$ && $21.5$ & $22.6$ & $16.8$ & $16.6\pm6.2$ \\ 
    & SDTN\cite{liu2020on} 
        &$14.0$& $14.6$ & $13.6$ & $18.6$ & $18.1$ & $8.1$ & $13.4$ && $10.3$ & $9.2$ & $17.2$ && $27.0$ & $35.5$ & $11.2$ & $16.2\pm7.6$ \\ 
    & CDCN\cite{cdcn}
        &$9.3$& $8.5$ & $13.9$ & $10.9$ & $21.0$ & $\mathbf{3.1}$ & $7.0$ && $45.0$ & $2.3$ & $16.2$ && $26.4$ & $20.9$ & $5.4$ & $14.6\pm11.7$ \\ 
    & CDCN++\cite{cdcn}
        &$12.4$& $8.5$ & $14.0$ & $13.2$ & $19.4$ & $7.0$ & $6.0$ && $45.0$ & $1.6$ & $14.0$ && $24.8$ & $20.9$ & $3.9$ & $14.6\pm11.4$ \\ 
    & HMP\cite{hmp2020}
        &$13.2$& $6.2$ & $13.1$ & $10.8$ & $16.3$ & $3.9$ & $\mathbf{2.3}$ && $34.1$ & $\mathbf{1.6}$ & $13.9$ && $23.2$ & $17.1$ & $2.3$ & $12.2\pm 9.4$ \\ 
    & Auxiliary\cite{liu-auxiliary-fas} 
        &$10.1$& $6.5$ & $10.9$ & $11.6$ & $\mathbf{6.2}$ & $7.8$ & $9.3$ && $11.6$ & $9.3$ & $7.1$ && $\mathbf{6.2}$ & $8.8$ & $10.3$ & $8.9\pm2.0$  \\ 
    & Ours
        &$\mathbf{3.8}$& $\mathbf{6.3}$ & $\mathbf{4.4}$ & $\mathbf{5.5}$ & $11.3$ & $3.5$ & $6.0$ && $\mathbf{6.6}$ & $1.8$ & $\mathbf{2.7}$ && $6.5$ & $\mathbf{8.0}$ & $\mathbf{1.1}$ & $\mathbf{5.7}\pm\mathbf{2.8}$ \\\hline
    \multirow{8}{*}{ACER} & LBP+SVM~\cite{OULU_NPU_2017}
        &$20.6$& $18.4$ & $31.3$ & $21.4$ & $45.5$ & $11.6$ & $13.8$ && $59.3$ & $23.9$ & $16.7$ && $35.9$ & $39.2$ & $11.7$ & $26.9\pm14.5$  \\ 
    & Auxiliary\cite{liu-auxiliary-fas}
        &$16.8$& $6.9$ & $19.3$ & $14.9$ & $52.1$ & $8.0$ & $12.8$ && $55.8$ & $13.7$ & $\mathbf{11.7}$ && $49.0$ & $40.5$ & $5.3$ &$23.6\pm18.5$ \\ 
    & DTL\cite{deep-tree}
        &$9.8$& $6.0$ & $15.0$ & $18.7$ & $36.0$ & $4.5$ & $13.4$ && $48.1$ & $11.4$ & $14.2$ && $19.3$ & $19.8$ & $8.5$ & $16.8\pm11.1$  \\ 
    & CDCN\cite{cdcn}
        &$8.7$& $7.7$ & $11.1$ & $\mathbf{9.1}$ & $20.7$ & $4.5$ & $5.9$ && $44.2$ & $2.0$ & $15.1$ && $25.4$ & $19.6$ & $3.3$ & $13.6\pm11.7$ \\ 
    & SDTN\cite{liu2020on} 
        &$7.8$& $7.3$ & $7.1$ & $12.9$ & $13.9$ & $4.3$ & $6.7$ && $53.2$ & $4.6$ & $19.5$ && $20.7$ & $21.0$ & $5.6$ & $14.2\pm13.2$  \\ 
    & CDCN++ \cite{cdcn}
        &$10.8$& $7.3$ & $9.1$ & $10.3$ & $18.8$ & $3.5$ & $5.6$ && $42.1$ & $0.8$ & $14.0$ && $24.0$ & $17.6$ & $1.9$ & $12.7\pm11.2$ \\ 
    & HMP\cite{hmp2020}
        &$12.8$& $5.7$ & $10.7$ & $10.3$ & $14.9$ & $\mathbf{1.9}$ & $\mathbf{2.4}$ && $\mathbf{32.3}$ & $\mathbf{0.8}$ & $\mathbf{12.9}$ && $22.9$ & $16.5$ & $1.7$ & $\mathbf{11.2}\pm\mathbf{9.2}$ \\ 
    & Ours
        &$\mathbf{6.9}$& $\mathbf{5.6}$ & $\mathbf{4.8}$ & $11.1$ & $\mathbf{7.4}$ & $2.7$ & $4.4$ && $49.7$ & $0.9$ & $20.1$ && $\mathbf{20.1}$ & $\mathbf{15.6}$ & $\mathbf{0.6}$ & $11.5\pm 13.2$ \\\hline
    \multirow{8}{*}{EER} & LBP+SVM~\cite{OULU_NPU_2017}
        &$20.8$& $18.6$ & $36.3$ & $21.4$ & $37.2$ & $7.5$ & $14.1$ && $51.2$ & $19.8$ & $16.1$ && $34.4$ & $33.0$ & $7.9$ & $24.5\pm12.9$ \\ 
    & Auxiliary\cite{liu-auxiliary-fas}
        &$14.0$& $4.3$ & $11.6$ & $12.4$ & $24.6$ & $7.8$ & $10.0$ && $72.3$ & $10.1$ & $\mathbf{9.4}$ && $21.4$ & $18.6$ & $4.0$  & $17.0\pm17.7$ \\ 
    & DTL\cite{deep-tree}
        &$10.0$& $\mathbf{2.1}$ & $14.4$ & $18.6$ & $26.5$ & $5.7$ & $9.6$ && $50.2$ & $10.1$ & $13.2$ && $19.8$ & $20.5$ & $8.8$ & $16.1\pm12.2$ \\ 
    & CDCN\cite{cdcn}
        &$8.2$& $7.8$ & $8.3$ & $\mathbf{7.4}$ & $20.5$ & $5.9$ & $5.0$ && $47.8$ & $1.6$ & $14.0$ && $24.5$ & $18.3$ & $1.1$ & $13.1\pm12.6$ \\ 
    & SDTN\cite{liu2020on}
        &$7.6$& $3.8$ & $8.4$ & $13.8$ & $14.5$ & $5.3$ & $4.4$ && $35.4$ & $0.0$ & $19.3$ && $21.0$ & $20.8$ & $1.6$ & $12.0\pm10.0$ \\
    & CDCN++\cite{cdcn}
        &$9.2$& $5.6$ & $\mathbf{4.2}$ & $11.1$ & $19.3$ & $5.9$ & $5.0$ && $43.5$ & $0.0$ & $14.0$ && $23.3$ & $14.3$ & $\mathbf{0.0}$ & $11.9\pm11.8$ \\ 
    & HMP\cite{hmp2020}
        &$13.4$& $5.2$ & $8.3$ & $9.7$ & $13.6$ & $5.8$ & $\mathbf{2.5}$ && $\mathbf{33.8}$ & $0.0$ & $14.0$ && $23.3$ & $16.6$ & $1.2$ & $11.3\pm 9.5$ \\ 
    & Ours
        &$\mathbf{5.2}$& $4.4$ & $4.4$ & $10.1$ & $\mathbf{8.6}$ & $\mathbf{2.6}$ & $4.3$ && $47.2$ & $\mathbf{0.0}$ & $19.6$ && $\mathbf{18.6}$ & $\mathbf{12.4}$ & $0.7$ & $\mathbf{10.6}\pm\mathbf{12.6}$ \\ \bottomrule
    TPR@ & SDTN\cite{liu2020on}
        &$45.0$& $40.5$ & $45.7$ & $36.7$ & $11.7$ & $40.9$ & $74.0$ && $0.0$ & $67.5$ & $16.0$ && $13.4$ & $9.4$ & $62.8$ & $35.7\pm23.9$ \\
    FNR=$.5\%$ & Ours
        &$\mathbf{55.1}$& $\mathbf{46.4}$ & $\mathbf{57.3}$ & $\mathbf{65.1}$ & $\mathbf{33.0}$ & $\mathbf{91.7}$ & $\mathbf{76.7}$ && $\mathbf{0.0}$ & $\mathbf{100.0}$ & $\mathbf{46.4}$ && $\mathbf{31.8}$ & $\mathbf{15.4}$ & $\mathbf{97.7}$ & $\mathbf{53.7}\pm\mathbf{31.8}$ \\ \bottomrule
	\end{tabular}
\label{tab:siwm_p2}} 
\end{table*}


%% file: sec-4-3-unknown.tex
\input{figures/table-siwm-p3}
\input{figures/figure10}
\SubSection{Anti-Spoofing for Unknown and Open-set Spoofs}
Another important aspect is to test the anti-spoofing performance on unknown spoof.
To use SiW-M for unknown spoof detection, The work \cite{deep-tree} defines the leave-one-out testing protocols, termed as SiW-M Protocol II. In this protocol, each model (\textit{i.e.}, one column in Tab.~\ref{tab:siwm_p2}) is trained with $12$ types of spoof attacks (as known attacks) plus the $80\%$ of the live faces, and tested on the remaining $1$ attack (as unknown attack) plus the $20\%$ of live faces.
As shown in Tab.~\ref{tab:siwm_p2}, our PhySTD achieves significant improvement over our preliminary version, with relatively $11.7\%$ on the overall EER, $19.0\%$ on the overall ACER, $50.4\%$ on the overall TPR. 
Specifically, we reduce the EERs of half mask, paper glasses, transparent mask, replay attack, and partial paper relatively by $47.6\%$, $40.4\%$, $37.7\%$, $31.6\%$, $56.3\%$, respectively.
Overall, compared with the top $7$ performances, we outperform the SOTA performance of EER/TPR and achieve comparable ACER.
Among all, the detection of silicone mask, paper-crafted mask, mannequin head, impersonation makeup, and partial paper attacks are relatively good, with the detection accuracy (\textit{i.e.}, TPR@FNR=$0.5\%$) above $65\%$. Obfuscation makeup is the most challenging one with TPR of $0\%$, where we predict all the spoof samples as live. This is due to the fact that the makeup looks very similar to the live faces, while being dissimilar to any other spoof types. However, once we obtain a few samples, our model can quickly recognize the spoof traces on the eyebrow and cheek, synthesize new spoof samples, and successfully detect the attack (TPR=$41.1 \%$ in Tab.~\ref{tab:siwm_p1}).
\input{figures/figure6}

Moreover, in the real-world scenario, the testing samples can be either a known spoof attack or an unknown one. Thus, we propose SiW-M Protocol III to evaluate this open-set testing situation. In Protocol III, we first follow the train/test split from protocol I, and then further remove one spoof type as the unknown attack. During the testing, we test on the entire unknown spoof samples as well the test split set of the know spoof samples. The results are reported in Tab.~\ref{tab:siwm_p3}. 
Compared to the SOTA face anti-spoofing method~\cite{liu-auxiliary-fas}, our approach substantially outperforms it in all three metrics.
\input{figures/table-cls}
\input{figures/figure7}

%% file: figures/table-siwm-p3.tex
\begin{table*}[t]
\caption{\small The evaluation on SiW-M Protocol III: openset spoof detection.}
\centering
	\resizebox{\textwidth}{!}{
	\begin{tabular}{rllllllllllllllllll}
	\multicolumn{17}{c}{} \\ \toprule
	Metrics & \multirow{2}{*}{Method} &\multirow{2}{*}{Replay}& \multirow{2}{*}{Print} & \multicolumn{5}{c}{3D Mask}  && \multicolumn{3}{c}{Makeup}  &&  \multicolumn{3}{c}{Partial Attacks} & \multirow{2}{*}{Overall}\\ 
	\cline{5-9} \cline{11-13} \cline{15-17}
	(\%)& & & & Half & Silic. & Trans. & Paper & Mann. && Ob. & Im. & Cos. && Funny. & Papergls. & Paper &\\ \hline
    \multirow{2}{*}{ACER} & Auxiliary\cite{liu-auxiliary-fas}
        &$6.7$& $5.6$ & $8.5$ & $7.5$ & $11.6$ & $6.7$ & $6.4$ && $8.9$ & $5.7$ & $6.1$ && $14.3$ & $15.9$ & $5.4$ & $8.4\pm3.4$ \\
    & Ours
        &$\textbf{4.7}$& $\textbf{3.5}$ & $\textbf{3.4}$ & $\textbf{3.3}$ & $\textbf{6.4}$ & $\textbf{2.6}$ & $\textbf{3.8}$ && $\textbf{7.0}$ & $\textbf{2.3}$ & $\textbf{3.2}$ && $\textbf{10.7}$ & $\textbf{7.3}$ & $\textbf{3.2}$ & $\textbf{4.7}\pm\textbf{2.4}$ \\\hline
    \multirow{2}{*}{EER} & Auxiliary\cite{liu-auxiliary-fas}
        &$6.4$& $5.6$ & $7.7$ & $6.5$ & $10.3$ & $6.1$ & $6.1$ && $8.4$ & $5.1$ & $6.3$ && $15.3$ & $13.1$ & $5.7$ & $7.9\pm3.2$ \\
    & Ours
        &$\textbf{4.1}$& $\textbf{2.8}$ & $\textbf{3.4}$ & $\textbf{3.1}$ & $\textbf{5.6}$ & $\textbf{3.6}$ & $\textbf{3.0}$ && $\textbf{6.7}$ & $\textbf{2.2}$ & $\textbf{3.4}$ && $\textbf{10.2}$ & $\textbf{8.6}$ & $\textbf{2.2}$ & $\textbf{4.5}\pm\textbf{2.5}$ \\ \hline
    TPR@ & Auxiliary~\cite{liu-auxiliary-fas}
        &$60.4$& $65.5$ & $64.4$ & $70.4$ & $47.5$ & $67.0$ & $71.6$ && $64.3$ & $75.1$ & $69.8$ && $45.8$ & $47.8$ & $62.9$ & $62.5\pm9.7$ \\
    FNR=$.5\%$& Ours
        &$\textbf{87.4}$& $\textbf{78.7}$ & $\textbf{81.0}$ & $\textbf{84.5}$ & $\textbf{69.0}$ & $\textbf{86.3}$ & $\textbf{84.7}$ && $\textbf{85.0}$ & $\textbf{91.0}$ & $\textbf{89.3}$ && $\textbf{66.6}$ & $\textbf{64.4}$ & $\textbf{91.1}$ & $\textbf{81.6}\pm\textbf{9.2}$ \\ \bottomrule
	\end{tabular}
\label{tab:siwm_p3}} 
\end{table*}

%% file: figures/figure10.tex
\begin{figure*}[t]
\small
\centering
\resizebox{1\linewidth}{!}{\includegraphics{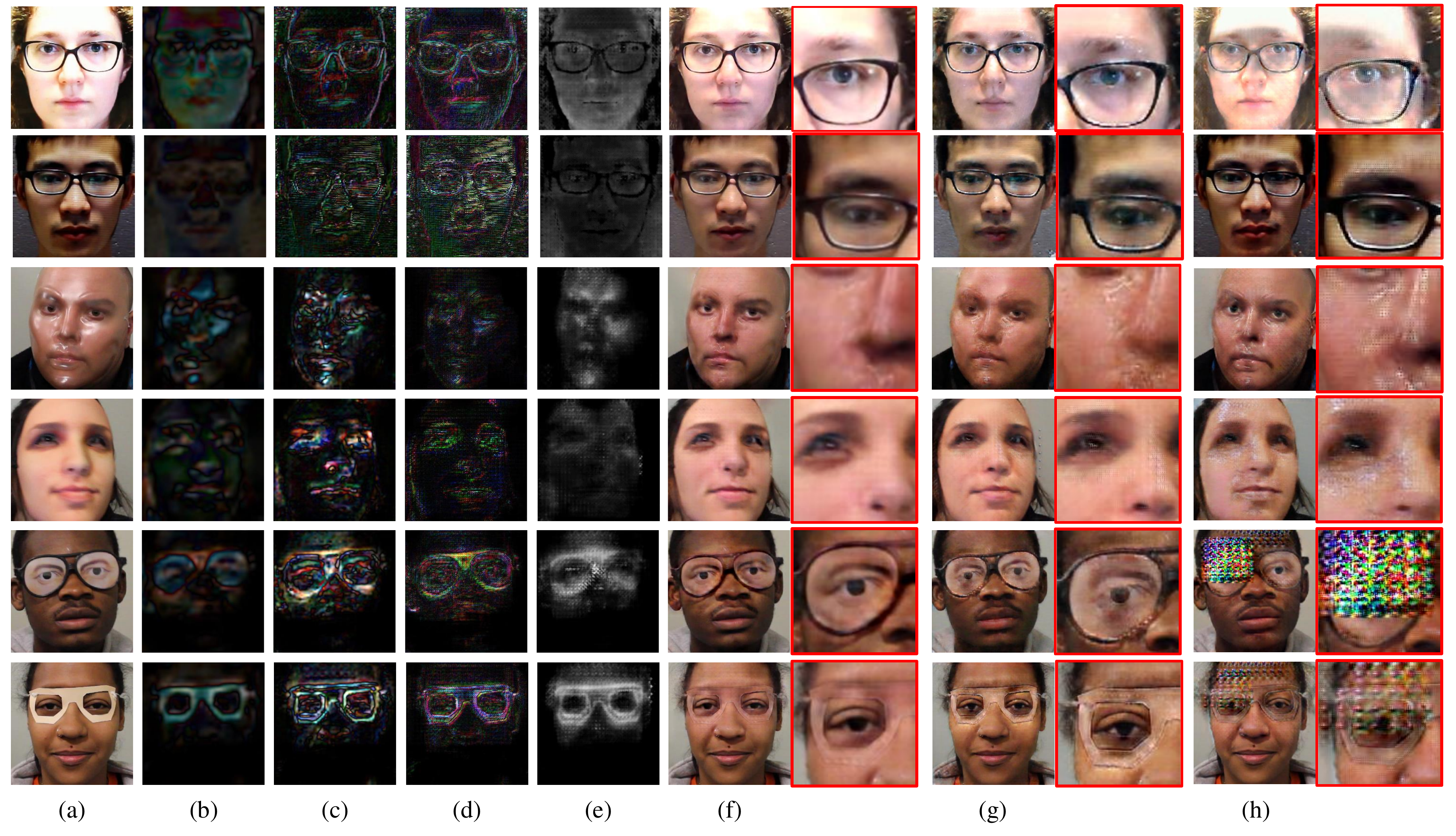}}
\caption{\small Examples of each spoof trace components. (a) the input sample faces. (b) $\textbf{B}$. (c) $\textbf{C}$. (d) $\textbf{T}$. (e) $\textbf{P}$. (f) the final live counterpart reconstruction and zoom-in details. (g) results from~\cite{liu2020on}. (h) results from Step$1$+Step$2$ with a single trace representation.
}
\label{fig:10}
\end{figure*}

%% file: figures/figure6.tex
\begin{figure*}[t]
\small
\centering
\resizebox{1\linewidth}{!}{\includegraphics{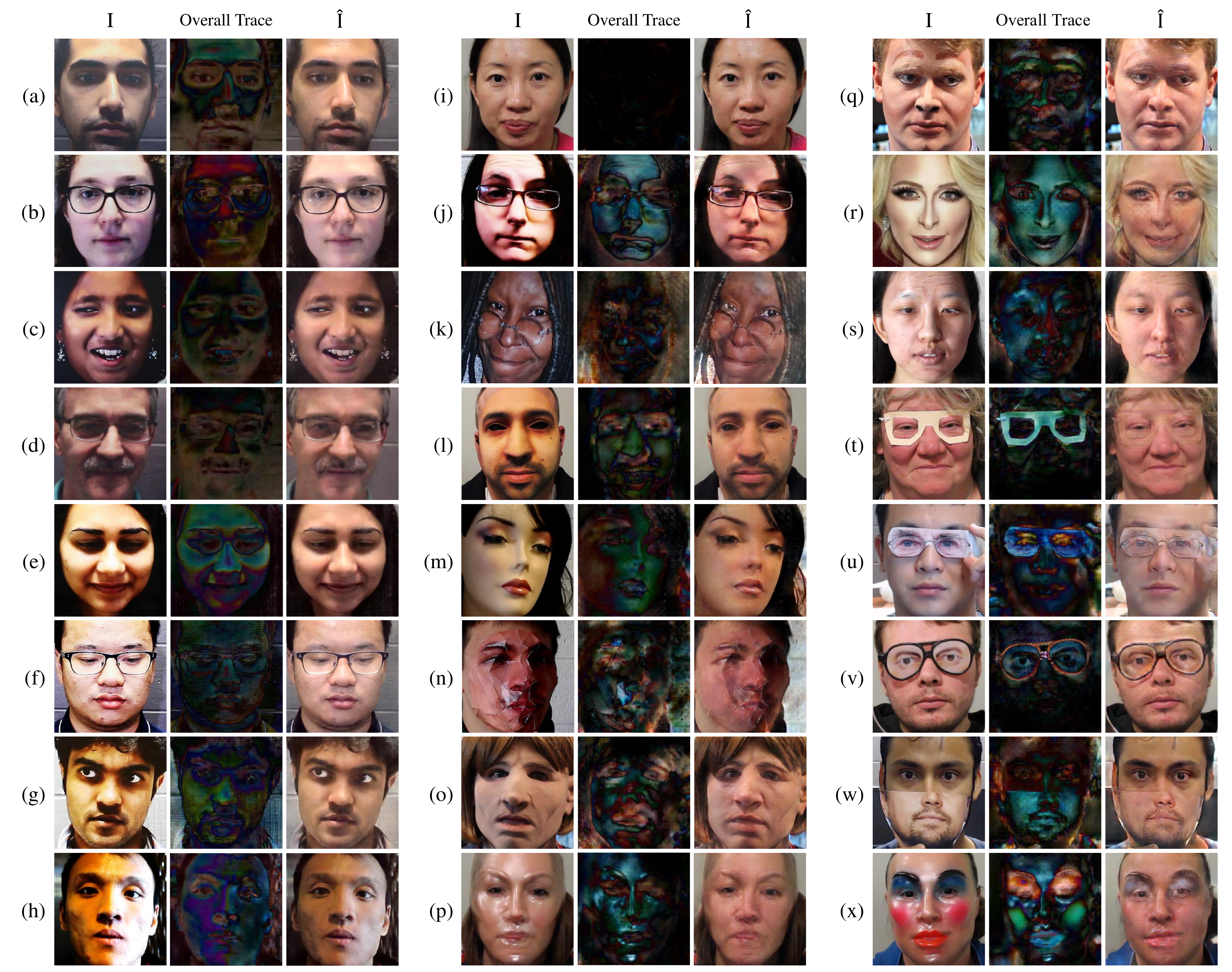}}
\caption{\small Examples of spoof trace disentanglement on SiW (a-h) and SiW-M (i-x). (a)-(d) items are print attacks and (e)-(h) items are replay attacks. (i)-(x) items are live, print, replay, half mask, silicone mask, paper mask, transparent mask, obfuscation makeup, impersonation makeup, cosmetic makeup, paper glasses, partial paper, funny eye glasses, and mannequin head. The first column is the input face, the second column is the overall spoof trace ($\textbf{I}-\hat{\textbf{I}}$), the third column is the reconstructed live.
}
\label{fig:6}
\end{figure*}

%% file: figures/table-cls.tex
\begin{table}[t!]
\small
\centering
\caption{\small Confusion matrices of spoof mediums classification based on spoof traces. The results are compared with the previous method~\cite{jourabloo-face-despoofing}. \textcolor{ForestGreen}{Green} represents improvement over~\cite{jourabloo-face-despoofing}.  \textcolor{red}{Red} represents performance drop.}
\resizebox{0.95\textwidth}{!}{
\begin{tabular}{llllll}
		\toprule
		\backslashbox{Label}{Predict}& Live & Print$1$ & Print$2$ & Replay$1$ & Replay$2$\\ \midrule
		Live       & \cellcolor{blue!12}$56(\color{red}{-4}$$)$ & $1(\color{red}{+1}$$)$  & $1(\color{red}{+1}$$)$  & $1(\color{red}{+1}$$)$  & $1(\color{red}{+1}$$)$ \\ 
		Print$1$   & $0$  & \cellcolor{blue!12}$43(\textcolor{ForestGreen}{+2})$ & $11(\textcolor{red}{+9})$  & $3(\textcolor{ForestGreen}{-8})$ & $3(\textcolor{ForestGreen}{-3})$ \\ 
		Print$2$   & $0$  & $9(\textcolor{ForestGreen}{-25})$ & \cellcolor{blue!12}$48(\textcolor{ForestGreen}{+37})$ & $1(\textcolor{ForestGreen}{-8})$  & $2(\textcolor{ForestGreen}{-4})$ \\ 
		Replay$1$ & $1(\textcolor{ForestGreen}{-9})$ & $2(\textcolor{ForestGreen}{-1})$  & $3(\textcolor{red}{+3})$  & \cellcolor{blue!12}$51(\textcolor{ForestGreen}{+38})$ & $3(\textcolor{ForestGreen}{-28})$ \\ 
		Replay$2$ & $1(\textcolor{ForestGreen}{-7})$  & $2(\textcolor{ForestGreen}{-5})$  & $2(\textcolor{red}{+2})$  & $3(\textcolor{ForestGreen}{-3})$  & \cellcolor{blue!12}$52(\textcolor{ForestGreen}{+13})$ \\ \bottomrule
	\end{tabular}}
\label{tab:cls}
\end{table}

\begin{table}[t!]
\small
\centering
\caption{\small Confusion matrices of $6$-class spoof traces classification on SiW-M database.}
\resizebox{0.95\textwidth}{!}{
\begin{tabular}{lllllll}
\toprule
\backslashbox{Label}{Predict}& Live & Print & Replay & Masks & Makeup & Partial\\ \midrule
Live       & \cellcolor{blue!12}$116$ & $6$  & $6$  & $3$  & $0$ & $0$\\ 
Print      & $1$  & \cellcolor{blue!12}$40$ & $1$  & $3$ & $0$ & $1$\\ 
Replay     & $3$  & $1$ & \cellcolor{blue!12}$32$ & $1$  & $0$ & $1$\\ 
Masks      & $3$ & $1$  & $1$  & \cellcolor{blue!12}$90$ & $0$ & $3$\\ 
Makeup     & $3$ & $0$  & $0$  & $0$ & \cellcolor{blue!12}$36$ & $0$ \\
Partial    & $2$  & $0$  & $0$  & $2$  & $0$ & \cellcolor{blue!12}$146$ \\ \bottomrule
\end{tabular}}
\label{tab:cls2}
\end{table}

%% file: figures/figure7.tex
\begin{figure*}[t]
\centering
\resizebox{\linewidth}{!}{\includegraphics{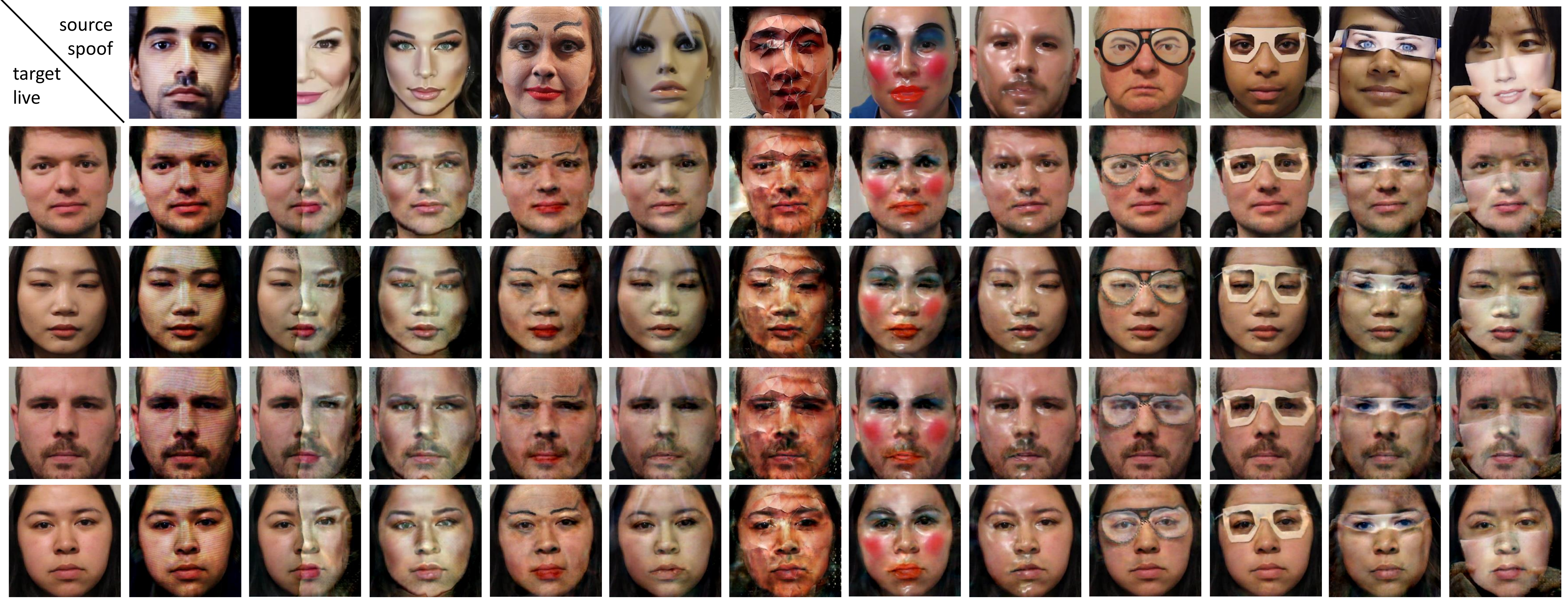}}
\caption{\small Examples of the spoof data synthesis. The first row are the source spoof faces, the first column are the target live faces, and the remaining are the synthesized spoof faces from the live face with the corresponding spoof traces.
}
\label{fig:7}
\end{figure*}

%% file: sec-4-4-trace-cls.tex
\SubSection{Spoof Traces Classification}
To quantitatively evaluate the spoof trace disentanglement, we perform a spoof medium classification on the disentangled spoof traces and report the classification accuracy.
The spoof traces should contain spoof medium-specific information, so that they can be used for clustering without seeing the face.
To make a fair comparison with ~\cite{jourabloo-face-despoofing}, we remove the additional spoof type information from the preliminary mask $\mathbf{P}_0$. That is, for this specific experiment, we only use the additive traces $\{\mathbf{B},\mathbf{C},\mathbf{T}\}$ to learn the trace classification.
After $\{\mathbf{B},\mathbf{C},\mathbf{T}\}$ finish training with only binary labels, 
we fix PhySTD and apply a simple CNN (\textit{i.e.}, AlexNet) on the estimated additive traces to do a supervised spoof medium classification.
We follow the same $5$-class testing protocol in~\cite{jourabloo-face-despoofing} in Oulu-NPU Protocol $1$.
We report the classification accuracy as the ratio between correctly predicted samples from all classes and all testing samples.
Shown in Tab.~\ref{tab:cls}. Our model can achieve a $5$-class classification accuracy of $83.3\%$. If we treat two print attacks as the same class and two replay as the same class,  our model can achieve a $3$-class classification accuracy of $92.0\%$.
Compared with the prior method~\cite{jourabloo-face-despoofing}, we show an improvement of $29\%$ on the $5$-class model.
In addition, we train the same CNN on the original images instead of the estimated spoof traces for the same spoof medium classification task, and the classification accuracy can only reach $80.6\%$.
This further demonstrates that the estimated traces do contain significant information to distinguish different spoof mediums.

We also execute the spoof traces classification task on more spoof types in SiW-M database. We leverage the train/test split on SiW-M Protocol $1$. We first train the PhySTD till convergence, and use the estimated traces from the training set to train the trace classification network. We explore the $6$-class scenario, shown in Tab.~\ref{tab:cls2}.
Our $6$-class model can achieve the classification accuracy of $92.0\%$. Since the traces are more distinct among different spoof types, this performance is even better than $5$-class classification on print/replay scenario in Oulu-NPU Protocol $1$. 
This further demonstrates that PhySTD can estimate spoof traces that contain significant information of spoof mediums and can be applied to multiple spoof types.

%% file: sec-4-5-ablation.tex
\SubSection{Ablation Study}
In this section, we show the importance of each design of our proposed approach on the SiW-M Protocol I, in Tab.\ref{tab:siwm_p1}. 
Our baseline is the auxiliary FAS~\cite{liu-auxiliary-fas}, without the temporal module. 
It consists of the backbone encoder and depth estimation network.
When including the image decomposition, the baseline becomes the training step $1$ in Alg.~\ref{alg:1}, as the traces are not activated without the training step $2$.
To validate the effectiveness of GAN training, we report the results from the baseline model with our GAN design, denoted as Step$1$+Step$2$. 
We also provide the control experiment where the traces are represented by a single component to demonstrate the effectiveness of the proposed $5$-element trace representation. This model is denoted as Step$1$+Step$2$ with single trace.
In addition, we evaluate the effect of training with more synthesized data via enabling the training step $3$ as Step$1$+Step$2$+Step$3$, which is our final approach.

As shown in Tab.~\ref{tab:siwm_p1}, the baseline model (Auxiliary) can achieve a decent performance of EER $6.7\%$.
Adding image decomposition to the baseline (Step 1) can improve the EER from $6.7\%$ to $4.3\%$, 
but more live samples are predicted with higher scores, causing a worse ACER. 
Adding simple GAN design (Step$1$+Step$2$ with single trace) may lead to a similar EER performance of $5.8\%$, but based on the TPR ($59.3\%\rightarrow 74.8\%$) its practical performance may be improved.
With the proper physics-guided trace disentanglement, we can improve the EER to $2.8\%$ and TPR to $89.7\%$.
And our final design can achieve the performance of HTER $2.8\%$, EER $2.5\%$, and TPR $91.2\%$.
Compared with our preliminary version, the EER is improved by $47.9\%$, HTER is improved by $31.7\%$ and TPR is improved by $29.5\%$.
\input{figures/figure3-supp}
\input{figures/figure2-supp}

%% file: figures/figure3-supp.tex
\begin{figure*}[t]
\small
\centering
    \includegraphics[width=1\textwidth]{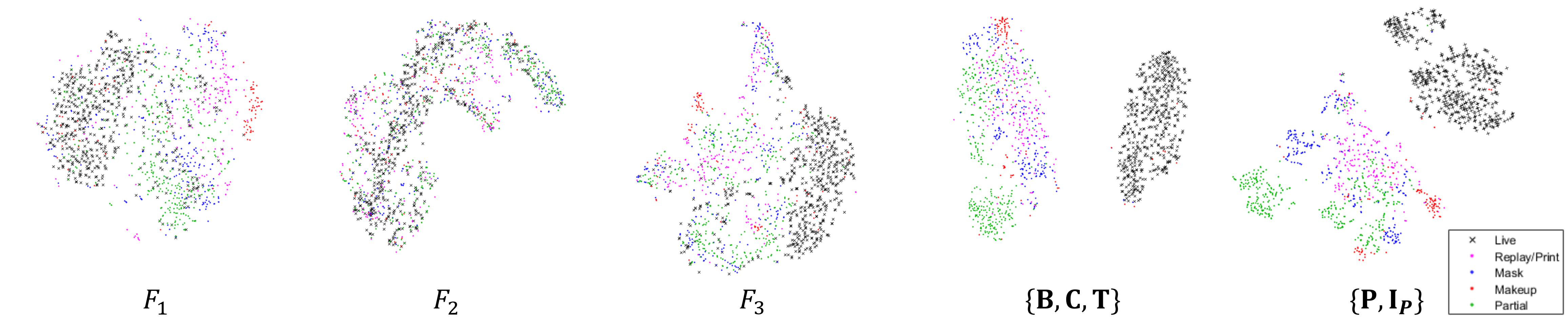}
    \captionof{figure}{\small The tSNE visualization of features from different scales and layers. The first $3$ visualization are from the encoder feature $F_1$,$F_2$,$F_3$, and the last $2$ visualization are from the features that produce $\{\textbf{B},\textbf{C},\textbf{T}\}$ and $\{\textbf{P},\textbf{I}_P\}$.}
    \label{fig:3-supp}
\end{figure*}

%% file: figures/figure2-supp.tex
\begin{figure}
\small
\centering
    \includegraphics[width=0.99\textwidth]{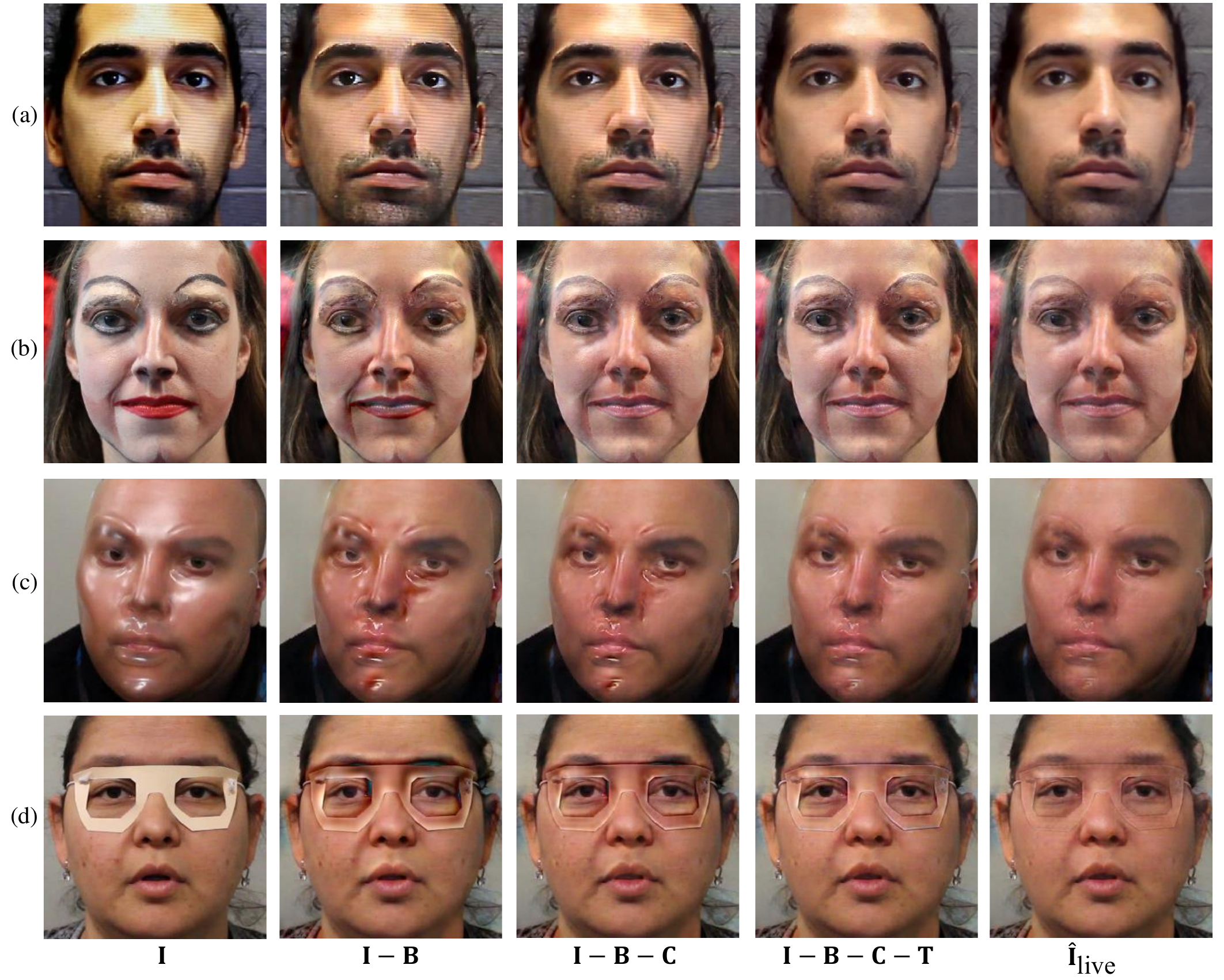}
    \captionof{figure}{\small The illustration of removing the disentangled spoof trace components one by one. The estimated spoof trace elements of input spoof (the first column) are progressively removed in the order of $\mathbf{\mathbf{B},\mathbf{C},\mathbf{T},\mathbf{T}_P}$. The last column shows the reconstructed live image after removing all three additive trace components and the inpainting trace. (a) Replay attack; (b) Makeup attack; (c) Mask attack; (d) Paper glasses attack.}
    \label{fig:2-supp}
\figvspace
\end{figure}%

%% file: sec-4-6-vis.tex
\SubSection{Visualization}
\Paragraph{Spoof trace components}
In Fig.\ref{fig:10}, we provide illustration of each spoof trace component.
Strong color distortion (low-frequency trace) shows up in the print attacks.
Moiré patterns in the replay attack are well detected in the high-frequency trace.
The local specular highlights in transparent mask are well presented in the low- and mid-frequency components, and the inpainting process further fine-tunes the most highlighted area.
For the two glasses attacks, the color discrepancy is corrected in the low-frequency trace, and the sharp edges are corrected in the mid- and high-frequency traces.
Each component shows a consistent semantic meaning on different spoof samples, and this successful trace disentanglement can lead to better final visual results.
As shown on the right side of Fig.~\ref{fig:10}, we compare with our preliminary version~\cite{liu2020on} and the ablated GAN design with a single trace representation.
The result of single trace representation shows strong artifacts on most of the live reconstruction.
The multi-scale from our preliminary version has already shown a large visual quality improvement, but still have some spoof traces (\textit{e.g.}, glass edges) remained in the live reconstruction. In contrast, our approach can further handle the missing traces and achieve better visualization.

\Paragraph{Live reconstruction}
In Fig.~\ref{fig:6}, we show more examples from different spoof types in SiW and SiW-M databases.
The overall trace is the exact difference between the input face and its live reconstruction. 
For the live faces, the trace is zero, and for the spoof faces, our method removes spoof traces without unnecessary changes, such as identity shift, and make them look like live faces.
For example, strong color distortion shows up in print/replay attacks (Fig.~\ref{fig:6}a-h) and some $3$D mask attacks (Fig.~\ref{fig:6}l-o).
For makeup attacks (Fig.~\ref{fig:6}q-s), the fake eyebrows, lipstick, artificial wax, and cheek shade are clearly detected.
The folds and edges (Fig.~\ref{fig:6}t-w) are well detected and removed in paper-crafted masks, paper glasses, and partial paper attacks.

\Paragraph{Spoof synthesis}
Additionally, we show examples of new spoof synthesis using the disentangled spoof traces, which is an important contribution of this work.
As shown in Fig.~\ref{fig:7}, the spoof traces can be precisely transferred to a new face without changing the identity of the target face. 
Due to the additional inpainting process, spoof attacks such as transparent mask and partial attacks can be better attached to the new live face.
Thanks to the proposed $3$D warping layer, the geometric discrepancy between the source spoof trace and the target face can be corrected during the synthesis.
Especially on the second source spoof, the right part of the traces is successfully transferred to the new live face while the left side remains to be still live.
It demonstrates that our trace regularization can suppress unnecessary artifacts generated by the network.
Both the live reconstruction results in Fig.~\ref{fig:6} and the spoof synthesis results in Fig.~\ref{fig:7} demonstrate that our approach disentangles visually convincing spoof traces that help face anti-spoofing.

\Paragraph{Spoof trace removing process}
As shown in Fig.~\ref{fig:2-supp}, we illustrate the effects of trace components by progressively removing them one by one. For the replay attack, the spoof sample comes with strong over-exposure as well as clear Moiré pattern. Removing the low-frequency trace can effectively correct the over-exposure and color distortion caused by the digital screen. And removing the texture pattern in the high-frequency trace can peel off the high-frequency grid effect and reconstruct the live counterpart. 

For the makeup attack, since there is no strong color range bias, removing estimated low-frequency trace would mainly remove the lip-stick color and fake eyebrow, but in the meantime bring a few artifacts at the edges.
Next, while removing the content pattern, the shadow on the cheek and the fake eyebrows are adequately lightened. 
Finally, removing the texture pattern would significantly correct the spoof traces from artificial wax, eyeliner, and shadow on the cheek.
Similarly, in mask and partial attacks, the reconstruction will be gradually refined as we removing components one by one.

\Paragraph{t-SNE visualization}
We use t-SNE~\cite{maaten2008visualizing} to visualize the encoder features $\mathbf{F}_1$,$\mathbf{F}_2$,$\mathbf{F}_3$, and the features that produce $\{\mathbf{B},\mathbf{C},\mathbf{T}\}$ and $\{\mathbf{P},\mathbf{I}_P\}$. 
The t-SNE is able to project the output of features from different scales and layers to $2$D by preserving the KL divergence distance. As shown in Fig.~\ref{fig:3-supp}, among the three feature scales in the encoder, $F_3$ is the most separable feature space, the next is $F_1$, and the worst is $F_2$. The features for additive traces $\{\textbf{B},\textbf{C},\textbf{T}\}$ are well-clustered as semantic sub-groups of live, makeup, mask, and partial attacks. 
As we know the inpainting masks for live samples are close to zero, the feature for inpainting traces $\{\textbf{P},\textbf{I}_P\}$ shows the inpainting process mostly update the partial attacks, and then some makeup attacks and mask attacks, \textit{i.e.}, the green dots being further away from the black dots means they have greater magnitude. 
This validates our prior knowledge of the inpainting process.

%% file: sec-5-0-main.tex
\Section{Conclusions}
This work proposes a physics-guided spoof traces disentanglement network (PhySTD) to tackle the challenging problem of disentangling spoof traces from the input faces.
With the spoof traces, we reconstruct the live faces as well as synthesize new spoofs.
To correct the geometric discrepancy in synthesis, we propose a $3$D warping layer to deform the traces.
The disentanglement not only improves the SOTA of face anti-spoofing in known, unknown, and open-set spoof settings, but also provides visual evidence to support the model's decision.

%% file: sec-6-0-ack.tex
\ifCLASSOPTIONcompsoc
  \section*{Acknowledgments}
\else
  \section*{Acknowledgment}
\fi

This research is based upon work supported by the Office of the Director of National Intelligence (ODNI), 
Intelligence Advanced Research Projects Activity (IARPA), via IARPA R$\&$D Contract No. 2017-17020200004. 
The views and conclusions contained herein are those of the authors and should not be interpreted as necessarily representing the official policies or endorsements, either expressed or implied, of the ODNI, IARPA, or the U.S. Government. The U.S. Government is authorized to reproduce and distribute reprints for Governmental purposes notwithstanding any copyright annotation thereon.

%% file: sec-7-0-ref.tex
\bibliographystyle{IEEEtran}
\bibliography{bib}

%% file: sec-8-0-authors.tex
\begin{IEEEbiography}[{\includegraphics[width=1in,height=1.2in,clip,keepaspectratio]{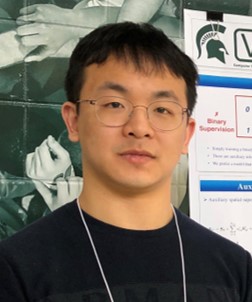}}]{Yaojie Liu}
is a PhD candidate at at Michigan State University in Computer Science and Engineering Department. He received his B.S. in Communication Engineering from University of Electronic
Science and Technology of China, and M.S. in Computer Science from the Ohio State
University. His research areas of interest are face representation and analysis, including face anti-spoofing, 2D/3D large pose face alignment, and 3D face reconstruction.
\end{IEEEbiography}

\begin{IEEEbiography}[{\includegraphics[width=1in,height=1.2in,clip,keepaspectratio]{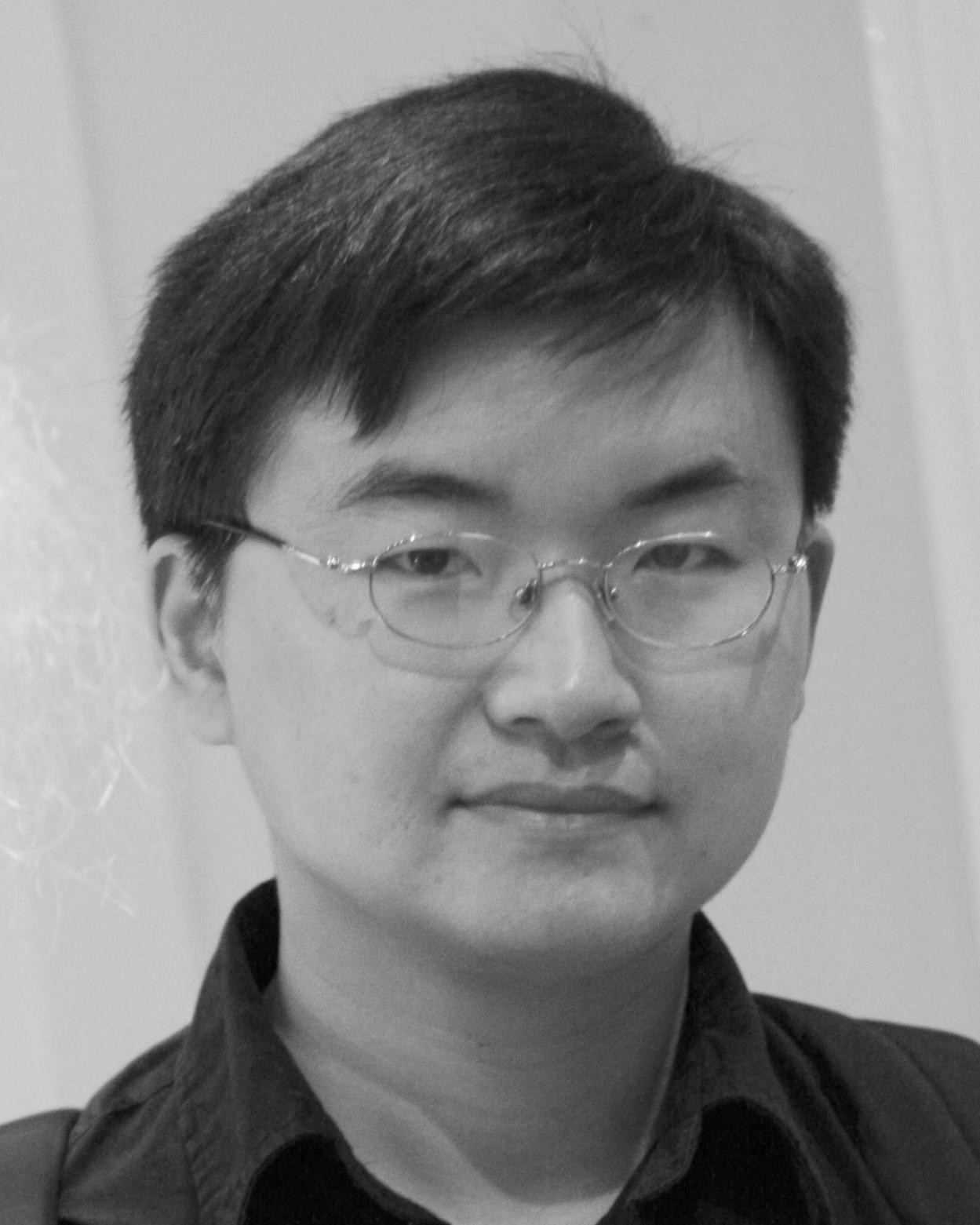}}]{Xiaoming Liu}
is a Professor at the Department of Computer Science and Engineering of Michigan State University. 
He received the Ph.D. degree in Electrical and Computer Engineering from Carnegie Mellon University in 2004.
Before joining MSU in Fall 2012, he was a research scientist at General Electric (GE) Global Research. 
His research interests include computer vision, machine learning, and biometrics.
As a co-author, he is a recipient of Best Industry Related Paper Award runner-up at ICPR 2014, Best Student Paper Award at WACV 2012 and 2014, Best Poster Award at BMVC 2015, and Michigan State University College of Engineering Withrow Endowed Distinguished Scholar Award. 
He has been Area Chairs for numerous conferences, including CVPR, ICCV, ECCV, ICLR, NeurIPS, the Program Co-Chair of BTAS’18, WACV’18, and AVSS’21 conferences, and General Co-Chair of FG’23 conference. He is an Associate Editor of Pattern Recognition Letters, Pattern Recognition, and IEEE Transaction on Image Processing. He has authored more than 150 scientific publications, and has filed 27 U.S.~patents. He is a fellow of IAPR.
\end{IEEEbiography}